\documentclass[11pt,a4paper]{article}
\usepackage[utf8]{inputenc}
\usepackage[T1]{fontenc}
\usepackage{mathptmx}
\usepackage[pdftex]{graphicx}
\usepackage[pdftex,linkcolor=black,pdfborder={0 0 0}]{hyperref}
\usepackage{calc}
\usepackage{enumitem}
\usepackage[all]{nowidow}
\usepackage[protrusion=true,expansion=true]{microtype}
\usepackage{listings}
\usepackage{amssymb}
\usepackage{amsmath}
\usepackage[a4paper, lmargin=0.14\paperwidth, rmargin=0.14\paperwidth, tmargin=0.111\paperheight, bmargin=0.1\paperheight]{geometry}
\usepackage{csquotes}
\usepackage[ backend=biber, style=numeric,sorting=none]{biblatex}
\usepackage{adjustbox}

\usepackage[protrusion=true,expansion=true]{microtype}
\usepackage{algorithm}
\usepackage{algorithmicx}
\usepackage{algpseudocode}
\usepackage{pgfplots}
\usepackage{pgfplotstable}
\pgfplotsset{compat=1.18}

\usepackage{tikz}
\usetikzlibrary{shapes.geometric, arrows, positioning, fit, backgrounds, calc, math, matrix ,positioning}

\hypersetup{ 	
    pdfsubject = {Master Thesis},
    pdftitle = {Vocabulary expansion through optimal initialization},
    pdfauthor = {Max Rehman Linder}
}

\begin{document}

% --------- FRONT PAGE ----------
\begin{titlepage}
    \centering
    \includegraphics[width=0.25\textwidth]{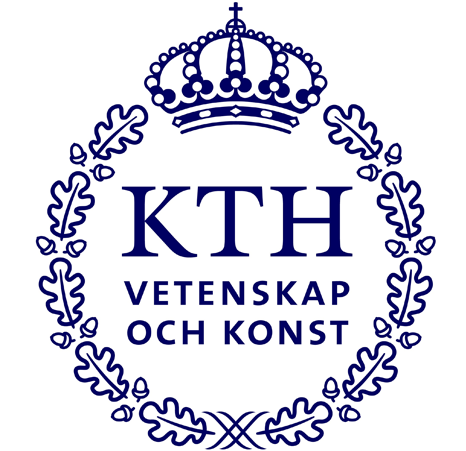}\par\vspace{1cm}
    {\scshape\LARGE KTH Royal Institute of Technology \par}
    \vspace{0.5cm}
    {\scshape\large School of Engineering Sciences \par}
    \vspace{0.5cm}
    {\scshape\large Master's Thesis in Applied Mathematics \par}
    \vspace{1.5cm}
    {\huge\bfseries Vocabulary Expansion of Large Language Models via Self-Distillation\par}
    \vspace{2cm}
    {\Large\itshape Max Rehman Linder\par}
    \vspace{0.5cm}
    {\large Degree Program in Engineering Physics\par}
    \vfill
    
    % Supervisors and date
    Supervisors: Lorenzo Vecchi \& Herman Forslund\\
    \vspace{0.3cm}
    Date: \today
    
    \vfill
\end{titlepage}

% ---------- TABLE OF CONTENTS -------------
\section*{Abstract}
Large pre-trained language models often struggle to incorporate new domain-specific terminology when fine-tuned on small, specialized corpora. In this work, the challenge of vocabulary expansion in frozen large language models is addressed by introducing a mathematically grounded method for knowledge distillation via Kullback–Leibler divergence, even for cases when the original and extended models use different tokenizations. This allows the student model to inherit distributional knowledge from the teacher despite differing vocabularies.

The Kullback-Leibler-based distillation approach is compared to conventional cross-entropy training, with both methods evaluated across multiple strategies for initializing new token embeddings. After embedding initialization, models are further fine-tuned to integrate the new vocabulary. Each trained model is benchmarked on approximately 2000 code-generation tasks, where the proposed approach achieves the best performance across the board.

Finally, through mechanistic interpretability, how models learn representations for the new tokens is analyzed, providing an explanation for the observed gains and offering insight into the structure of embedding space during vocabulary expansion.

\section*{Abstrakt}
Stora förtränade språkmodeller har ofta brister i att inkorporera ny domänspecifik terminologi och kunskap när de tränas på små, specialiserade korpusar. I detta arbete adresseras utmaningen med vokabulärexpansion i frysta språkmodeller genom att introducera en matematiskt grundad metod för kunskapsdestillation via Kullback-Leibler-divergens, även för fall när de ursprungliga och utökade modellerna använder någolunda olika tokeniseringar. Detta låter elevmodellen att ärva distributionell kunskap från lärarmodellen trots olika vokabulärer.

Kullback-Leibler-baserade destillationsmetoden jämförs med konventionell kors-entropi-träning, där båda metoderna utvärderas över flera strategier för initialisering av nya token-embeddingar. Efter embedding-initialisering finjusteras modellerna vidare för att integrera det nya vokabuläret. De tränade modellerna utvärderas på ungefär 2000 kodgenereringsuppgifter, där den föreslagna metoden uppnår bäst prestanda genomgående.

Slutligen analyseras, genom mekanistisk tolkbarhet, hur modeller lär sig representationer för de nya tokens, vilket ger en förklaring till de observerade förbättringarna och erbjuder insikt i strukturen av vektorrummet i vilket embeddings existerar under vokabulärexpansion.
\newpage
\tableofcontents
\newpage

% ---------- CHAPTERS ----------
\section{Introduction}

Large language models have transformed software engineering workflows, yet adapting these models to specialized domains remains a significant challenge. This work investigates vocabulary expansion as a mechanism for improving fine-tuning efficiency when introducing new programming languages or domain-specific notation to pretrained models. The following section provides a brief overview of the field, formalizes the problem being addressed, and outlines the proposed approach.

\subsection{Motivation}

In recent years, large language models (LLMs) have become a dominant focus of research and application, primarily due to their remarkable ability to augment engineers in solving novel problems and answering complex queries using their internal knowledge.

One field particularly affected by this development is programming, where engineers can become significantly more efficient by leveraging LLMs for writing code, debugging, and exploring alternative solutions. Trained on massive corpora, LLMs not only acquire general linguistic intelligence, but also develop a nuanced understanding of specific programming languages and conventions.

\vspace{0.5em}
The generative capabilities of LLMs stem largely from their exposure to a diverse and extensive set of training data. However, when adapting such models to new tasks or incorporating novel internal knowledge, the standard approach involves fine-tuning an open-source foundation model. This process allows the model to retain its understanding of natural language and general reasoning while learning new, task-specific behavior.

\vspace{0.5em}
A key challenge emerges in this adaptation: the training data for a new domain are typically much smaller and less diverse than the terabytes of information used during pre-training. This means that there lies a challenge in making the model proficient in the new programming language. This is especially true if the notation of that language differs from that of the code on which the model was trained.

\subsection{Research Question}

One avenue for improving the fine-tuning process is through vocabulary expansion. By extending the vocabulary, the number of tokens the model can represent directly is effectively increased. The purpose of this is to compress the training data more effectively and enable the model to learn common notations and variable names using denser, more meaningful representations. However, enabling the model to utilize new representations for input and output requires the introduction of new components into the model—most notably, new embeddings.

While it is generally assumed that larger models benefit from larger vocabularies~\cite{tao2024scalinglawsvocabularylarger}, in most cases, the tokenizer and model are trained jointly from scratch. In contrast, this investigation focuses on how the vocabulary of a pre-trained model can be retroactively extended and trained to integrate new tokens effectively.

The key research question is how these new components should be initialized using a novel student–teacher framework. Beyond simple random initialization or heuristically derived values (e.g., mean of existing embeddings), mathematically grounded approaches for optimizing the new embeddings are investigated. The goal is to make them fit as seamlessly as possible into the pretrained model's structure. In essence, it is aimed to determine how sequences of existing embeddings can be compressed into a single embedding that captures their contextual meaning, using a training objective that aligns the new embedding with the composition of the original ones.

This line of inquiry lies at the intersection of applied fine-tuning and foundational research on how large language models interpret and utilize embeddings. By testing different initialization strategies, including the optimized approach presented here, insight is gained into what characteristics of an embedding influence performance. Comparing these learned embeddings to heuristic baselines also allows for evaluation of to what extent common intuitions about embedding space and compositionality hold in practice.

\subsection{Methodology Outline}

A novel student-teacher framework is introduced where the pre-trained model serves as its own teacher for learning new token representations. The methodology consists of three main phases that systematically address vocabulary expansion challenges.

\vspace{0.5em}
\textbf{Embedding Initialization via Self-Distillation:} The model itself is leveraged as a teacher to initialize new embeddings. Rather than using heuristic approaches like averaging constituent token embeddings, new embeddings are optimized using Kullback–Leibler divergence loss. The training objective ensures that sequences tokenized with the extended vocabulary produce output distributions that closely match those from the original tokenization. This self-distillation approach allows new tokens to integrate seamlessly into the pretrained model's existing knowledge structure.

\vspace{0.5em}
\textbf{Head Extension with Strategic Initialization:} For the output head that predicts next tokens, the vocabulary is extended by copying existing token representations. Specifically, when a new token represents a composition of existing tokens (e.g., "numpy" from "num" + "py"), the new token's head parameters are initialized using the first constituent token's values. This provides a better starting point than random initialization for subsequent cross-entropy training.

\vspace{0.5em}
\textbf{Progressive Fine-tuning Pipeline:} The complete training process involves multiple phases: initial embedding and head optimization, full vocabulary adaptation training, and final domain-specific fine-tuning using LoRA on all transformer blocks. This progressive approach ensures that new tokens are properly integrated before full model adaptation.

\begin{figure}[htbp]
\begin{center}
\resizebox{0.99\linewidth}{!}{
\begin{tikzpicture}[
    box/.style={rectangle, draw, rounded corners, minimum width=2.5cm, minimum height=1cm, align=center, fill=blue!20},
    arrow/.style={->, >=stealth, thick}
]
    % Boxes
    \node[box] (pretrain) at (0,0) {Pre-Training\\General Knowledge};
    \node[box, fill=red!20] (vocab) at (3.1,0) {Vocabulary\\Expansion};
    \node[box] (code) at (6.05,0) {Code Completion\\Fine-tuning};
    \node[box] (instruction) at (9,0) {Instruction\\Tuning};
    
    % Arrows
    \draw[arrow] (pretrain.east) -- (vocab.west);
    \draw[arrow] (vocab.east) -- (code.west);
    \draw[arrow] (code.east) -- (instruction.west);
    
    % Labels
    \node[align=center, font=\footnotesize] at (0,-1.3) {Massive corpus\\General capabilities};
    \node[align=center, font=\footnotesize, text=red] at (3,-1.3) {Our focus\\New token learning};
    \node[align=center, font=\footnotesize] at (6,-1.3) {Domain-specific\\code knowledge};
    \node[align=center, font=\footnotesize] at (9,-1.3) {Follow user\\instructions};
\end{tikzpicture}
}
\end{center}
\caption{Complete pipeline of vocabulary expansion and subsequent finetuning of LLM}
\label{fig:complete_pipeline}
\end{figure}

Figure \ref{fig:complete_pipeline} highlights the process of training a large language model. It is noted that usually the addition of tokens is done before pre-training, whereas in this setting a trained model is started with and tokens are retroactively added. This means the model has to be taught to use the added tokens while also maintaining the language modeling capabilities learned during pre-training.

The approach is evaluated using standardized code generation benchmarks including BigCodeBench and DS-1000, comparing multiple initialization strategies to demonstrate the effectiveness of the proposed method. This investigation is particularly motivated by practical applications at Ericsson, where effective vocabulary expansion could enable better adaptation of LLMs to proprietary programming languages and domain-specific notation.

The following aspects are of particular interest:
\begin{itemize}
    \item How vocabulary extension can be performed effectively in a pretrained model,
    \item Performance on a range of standardized benchmarks,
    \item Understanding and interpretation of LLM behavior.
\end{itemize}
\section{Background}

This section provides a comprehensive background on the theory and engineering foundations upon which this thesis builds. The training of modern large AI models relies on a convergence of mathematics, computer science, and engineering, encompassing a wide array of domain-specific terms and architectures. The objective of this section is to establish the necessary theoretical framework and terminology required to fully understand the motivations and intuitions behind the experiments proposed in this thesis.

\subsection{Mathematical background for autoregressive models and deep learning}

This section introduces the fundamentals of Natural Language Processing (NLP), focusing on how text is transformed into learnable representations for modeling. It also covers the core principles of deep learning, which provides the mathematical framework for learning complex, non-linear mappings of arbitrary data distributions. These two essential fields form the backbone of Large Language Models (LLMs), enabling them to effectively model the inherent complexities of natural language.

\subsubsection{Autoregressive Language Modeling}

The primary goal of a Large Language Model (LLM) is to predict text that accurately emulates natural language or provides a useful, coherent response to a given query. The standard training objective is to maximize the likelihood that the model outputs the reference answer provided in the training data.

\begin{figure}[htbp]
    \centering
    \includegraphics[width=0.99\textwidth]{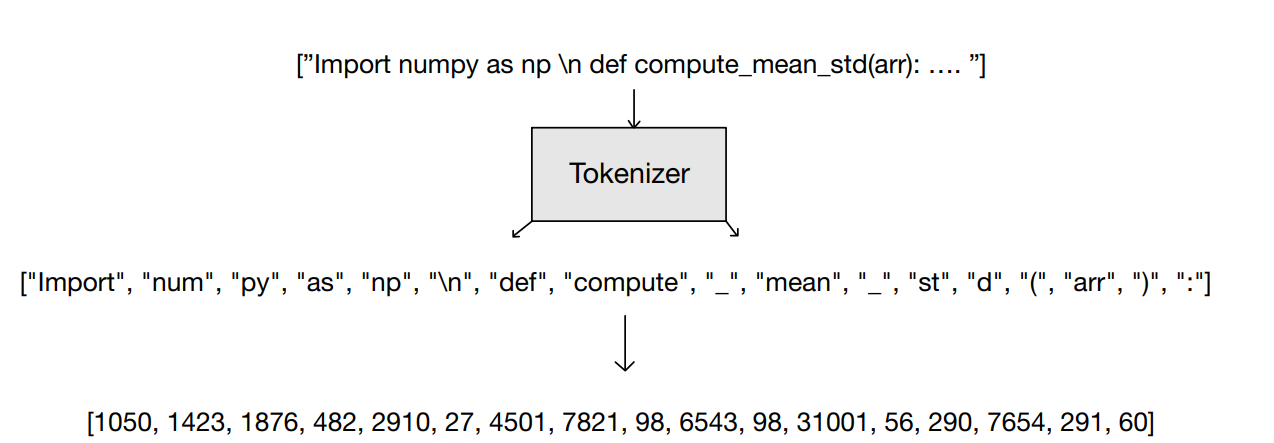}
    \caption{Visualisation of tokenization process.}
    \label{fig:token1}
\end{figure}

Given a dataset $D$ and a model with parameters $\theta$, the objective is to find parameter values $\theta^*$ that maximize the probability of generating the correct sequence of characters across the entire dataset. For language modeling, characters are chunked into \textbf{tokens}, where one token represents one or, most often, several characters as can be seen in Figure \ref{fig:token1}. In this thesis, a single token is denoted by $t$. Consequently, for a given sequence, the number of tokens is smaller than the number of characters. How tokens are constructed and why they are chosen over individual characters will be discussed in the next subsection. The key point to take away is that natural language, no matter how it is tokenized, can be viewed as a sequence of tokens that form a probability distribution.

The true distribution $p^*(t_{i+1} \mid t_{1:i})$ is denoted as the (unknown) true next-token distribution of the data-generating process, and $q_\theta(t_{i+1} \mid t_{1:i})$ as the distribution induced by the model. This relationship between dataset size, model capacity, and achievable performance is governed by scaling laws~\cite{kaplan2020scaling}, which have been a key driver in recent advancements in AI.

\begin{figure}[H]
    \centering
    \hspace{-0.5cm}
    \makebox[\textwidth]{\includegraphics[width=1.3\textwidth]{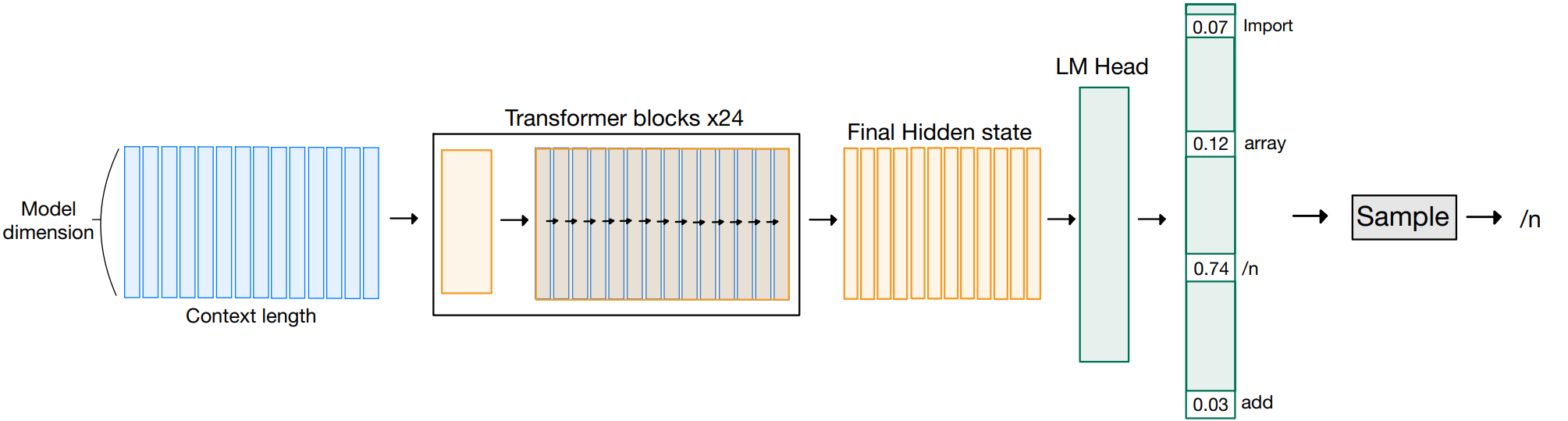}}
    \caption{Illustration of the pipeline of language model at inference, where we sample from the produce distribution.}
    \label{fig:llm_pipe}
\end{figure}

Mathematically, the model is represented as a function $f_\theta$ that maps a context $t_{1:i}$ (or its corresponding embedding sequence $\mathbf{t}_{1:i}$) to a probability distribution over the next token:
\begin{equation}
    q_\theta(t_{i+1} \mid t_{1:i}) \;=\; f_\theta(\mathbf{t}_{1:i}).
     \label{eq:model_likelihood}
\end{equation}
In other words, the probability distribution $q_\theta(\cdot \mid t_{1:i})$ of the subsequent token $t_{i+1}$ is predicted by the model, conditioned on the preceding input sequence as seen in FOgure \ref{fig:llm_pipe}

By construction, LLMs are \textbf{autoregressive}, meaning the output distribution at any index $i$ depends only on the tokens that preceded it. This property is also key to text generation: during inference, a newly generated token is immediately fed back into the model as part of the new input sequence to predict the next token.

\subsubsection{Probability and Loss Function}

Given that language modeling is about constructing a model that assigns a plausible probability distribution to the next token in a sequence, the problem needs to be formalized mathematically so that it can be optimized.

Let $D = \{\mathbf{t}^{(s)}_{1:L_s}\}_{s=1}^N$ denote the training corpus, where $\mathbf{t}^{(s)}_{1:L_s} = (t^{(s)}_1,\dots,t^{(s)}_{L_s})$ is the $s$-th sequence of tokens and $N$ is the number of sequences. An autoregressive model with parameters $\theta$ defines a conditional distribution $q_\theta(t_i \mid t_{<i})$ over the next token given the previous tokens $t_{<i} = (t_1,\dots,t_{i-1})$. By the chain rule, the likelihood of a single sequence is
\begin{equation}
    q_\theta\big(\mathbf{t}^{(s)}_{1:L_s}\big)
    = \prod_{i=1}^{L_s} q_\theta\!\big(t^{(s)}_i \mid t^{(s)}_{<i}\big),
    \label{eq:seq_likelihood}
\end{equation}
and the likelihood of the whole dataset is
\begin{equation}
    p(D \mid \theta)
    = \prod_{s=1}^N q_\theta\big(\mathbf{t}^{(s)}_{1:L_s}\big).
    \label{eq:data_likelihood}
\end{equation}
The corresponding log-likelihood is therefore
\begin{equation}
    \log p(D \mid \theta)
    = \sum_{s=1}^N \sum_{i=1}^{L_s}
      \log q_\theta\!\big(t^{(s)}_i \mid t^{(s)}_{<i}\big).
    \label{eq:log_likelihood}
\end{equation}

Maximizing \eqref{eq:log_likelihood} with respect to $\theta$ is equivalent to minimizing the average negative log-likelihood per token. Let $M = \sum_{s=1}^N L_s$ denote the total number of tokens in the corpus. The average loss per token is defined as
\begin{equation}
    \mathcal{L}(\theta)
    = -\frac{1}{M} \sum_{s=1}^N \sum_{i=1}^{L_s}
      \log q_\theta\!\big(t^{(s)}_i \mid t^{(s)}_{<i}\big).
    \label{eq:avg_token_loss}
\end{equation}
Isolating the inner term, it can be seen that for each position $i$ in each sequence $s$, the contribution to the loss is simply $-\log q_\theta(t^{(s)}_i \mid t^{(s)}_{<i})$. Put in words, the more abstract idea of maximizing the probability of generating the entire dataset has been moved to something more tangible: for each token, parameters for the model are sought such that, given the previous tokens, the negative log-likelihood of the observed next token is as small as possible on average. This is the loss function being minimized.

In this thesis, \eqref{eq:avg_token_loss} is referred to as the \emph{cross-entropy loss}.

However, while this is the actual quantity being minimized, it is worth addressing the formal definition of cross-entropy. For an individual context $t_{<i}$, the cross-entropy between a target distribution $p(\cdot \mid t_{<i})$ and the model distribution $q_\theta(\cdot \mid t_{<i})$ over the vocabulary $V$ (with $|V|$ possible tokens) is defined as
\begin{equation}
    H\big(p(\cdot \mid t_{<i}), q_\theta(\cdot \mid t_{<i})\big)
    = - \sum_{t \in V} p(t \mid t_{<i}) \log q_\theta(t \mid t_{<i}).
    \label{eq:cross_entropy_context}
\end{equation}
Here $V$ denotes the vocabulary used by the model, and $|V|$ is the size of this probability space of possible next tokens.

This definition follows the conventions from information theory, where the entropy of a distribution $p$ over $V$ is given by
\begin{equation}
    H(p) = - \sum_{t \in V} p(t) \log p(t),
\end{equation}
and the cross-entropy $H(p, q)$ represents the expected log-loss incurred when the data are generated from $p$ but predictions are made using $q$.

In the language-modeling setting, $p$ is typically identified with the empirical data distribution $p_{\text{data}}$. For each context $t_{<i}$ in the dataset, the conditional distribution $p_{\text{data}}(t_i \mid t_{<i})$ is represented as a one-hot vector: probability $1$ is assigned to the observed (target) token $t_i$ and probability $0$ to all other tokens in $V$. Substituting this into \eqref{eq:cross_entropy_context} shows that, for a single training example,
\[
    H\big(p_{\text{data}}(\cdot \mid t_{<i}), q_\theta(\cdot \mid t_{<i})\big)
    = - \log q_\theta(t_i \mid t_{<i}),
\]
so each per-token loss term is exactly the cross-entropy between the one-hot target distribution and the model's predicted distribution over the vocabulary. Averaging this quantity over all token positions in the corpus yields the average per-token loss defined in Equation \ref{eq:avg_token_loss}.

\subsubsection{Stochastic Gradient Descent}

Scaling laws for natural language models~\cite{kaplan2020scaling} indicate that increasing model size and data volume leads to consistently lower loss. Since natural language is inherently complex, Large Language Models (LLMs), in particular, contain vast numbers of parameters—often in the billions. Due to the sheer size and complexity of both the data distribution and the model, no analytically tractable way exists to find the optimal parameters $\theta^*$ that minimize Equation \ref{eq:avg_token_loss}.

For this reason, Stochastic Gradient Descent (SGD)~\cite{robbins1951stochastic}, or variations thereof such as AdamW~\cite{loshchilov2017decoupled}, is utilized by LLMs to iteratively optimize $\theta$. The procedure generally follows Algorithm \ref{alg:sgd}.

\begin{algorithm}[H]
\caption{Stochastic Gradient Descent Step}
\label{alg:sgd}
\begin{algorithmic}
\State \textbf{Input:} Dataset $D$, Batch size $B$, Learning rate $\eta$, Model parameters $\theta$
\For{each batch $b = \{(\mathbf{t}^{(j)}_{1:L}) \}_{j=1}^B$ in $D$}
    \State \textit{1. Forward Pass:}
    \State Calculate loss over the batch:
    \State $$ \mathcal{L}_{\text{batch}}(\theta) = -\frac{1}{B \times L} \sum_{j=1}^{B} \sum_{i=1}^{L} \log q_\theta(t^{(j)}_i \mid t^{(j)}_{<i}) $$
    
    \State \textit{2. Backward Pass:}
    \State Compute gradients using the chain rule (Backpropagation):
    \State $$ \nabla_\theta \mathcal{L} = \frac{\partial \mathcal{L}_{\text{batch}}}{\partial \theta} $$
    
    \State \textit{3. Parameter Update:}
    \State Update parameters in the direction of negative gradient:
    \State $$ \theta \leftarrow \theta - \eta \nabla_\theta \mathcal{L} $$
\EndFor
\end{algorithmic}
\end{algorithm}

While the update algorithm itself is deterministic, the process is termed ``stochastic'' because the order in which data samples are fed to the model is typically randomized, introducing noise into the gradient estimation that helps escape local minima.

A fundamental requirement for this method is that all computational steps involved in calculating the loss, including the loss function itself, must be differentiable to allow for gradient computation via the chain rule.

\begin{figure}[htbp]
    \centering
    \begin{tikzpicture}
    \begin{axis}[
        width=0.8\textwidth,
        height=6cm,
        xlabel={Predicted Probability ($p$)},
        ylabel={Value},
        xmin=0, xmax=1,
        ymin=0, ymax=5,
        grid=major,
        legend pos=north east,
        legend entries={Cross Entropy Loss ($-\log p$), Gradient Magnitude ($|1/p|$)}
    ]
        % Note: Ensure 'data/ce_loss_data.csv' exists with columns: probability, loss, gradient
        \addplot [blue, thick] table [x=probability, y=loss, col sep=comma] {data/ce_loss_data.csv};
        \addplot [red, dashed, thick] table [x=probability, y=gradient, col sep=comma] {data/ce_loss_data.csv};
    \end{axis}
    \end{tikzpicture}
    \caption{Loss value and corresponding gradient magnitude for the ground truth label, as a function of the predicted probability $p$.}
    \label{fig:loss_gradient}
\end{figure}

As illustrated in Figure \ref{fig:loss_gradient}, the loss approaches infinity if the model assigns a probability of 0 to the ground truth token, and approaches 0 as the probability approaches 1. Two key intuitions can be derived from this relationship.

Firstly, although the training data typically consists of one-hot encoded vectors (where the target is 100\% certain), a probability distribution is effectively learned by the model. Minimizing loss across the dataset is equivalent to finding a compromise for the model distribution $q_\theta$ such that no plausible continuation is assigned a probability near zero. If a plausible token were assigned a near-zero probability, the resulting loss would be excessively large. Consequently, the model would be heavily penalized by Stochastic Gradient Descent, forcing a sharp increase in that token's likelihood given the context.

Secondly, it is observed that even when the loss approaches 0 (as the predicted probability $p \to 1$), the gradient with respect to the probability remains non-zero. Specifically, for cross-entropy loss $\mathcal{L} = -\log(p)$, the gradient with respect to the output probability is $\frac{\partial \mathcal{L}}{\partial p} = -\frac{1}{p}$. As $p \to 1$, the magnitude of this gradient approaches 1.

This stands in contrast to Mean Squared Error (MSE), defined as $(1-p)^2$, where the gradient $-2(1-p)$ vanishes to 0 as the prediction approaches the target~\cite{golik2013cross}. This property of Cross Entropy is crucial for training stability and efficiency. The intuition is that even if the model accurately predicts a token, a persistent gradient exists that "pushes" the model to maintain that confidence. If a loss of 0 resulted in a gradient of 0 w.r.t the probability output, the model could stagnate or drift; the non-zero gradient ensures a robust learning signal is maintained throughout the optimization process.

\subsection{LLM Architecture}

While the previous subsection described an LLM from a purely mathematical and theoretical perspective, the following will delve into the core components of how an LLM is actually constructed. This section places particular emphasis on embeddings, as these are central to understanding the proposed method in this thesis. 

An important point to keep in mind is that all matrices denoted by symbols such as $\mathbf{W}$ are \emph{weight matrices}. Each entry in such a matrix is a trainable parameter. The sheer number and size of these matrices are what make modern Large Language Models so large in terms of parameter count. The illustrative figures with ``toy'' numerical values depict the resulting outputs after these various matrix multiplications have been applied.

Relating back to Algorithm~\ref{alg:sgd}, each parameter in these matrices is differentiable with respect to the loss. During the backward pass, gradients are propagated through all intermediate computations, and every entry of each weight matrix is updated according to the chosen optimization rule (e.g., SGD or AdamW).

\subsubsection{Tokens and Embeddings}

For a model to ``learn'' language and text, representations that are processable by neural networks must be utilized. While a text editor works with characters as bits (normally UTF-8) that map a binary code to a character, this is not optimal for a model. The main reason is that if a letter or word is assigned a hard-coded sequence of zeros and ones, there is no reason why words close in Hamming (binary) space would be close in semantic meaning, or easily distinguishable from one another.

Therefore, a sequence of text is transformed into a sequence of embeddings, where characters and/or words are mapped to a sequence of vectors. These vectors are typically high-dimensional, allowing them to encode information in a much more decompressed state than the binary form read as text. Crucially, they are trainable, meaning the entries of the vectors are unknown beforehand and are optimized during training like the rest of the model parameters.

Before embeddings are attributed to a sequence of text, the text is first tokenized. This means it is chunked into words or subwords to form a sequence of discrete units. Table \ref{tab:tokenization_process} illustrates this process: the raw code string is split into tokens, which are then mapped to specific integer indices (IDs) from the model's vocabulary.

\begin{table}[h]
\centering
\small
\setlength{\tabcolsep}{4pt}
\begin{tabular}{|l|c|c|c|c|c|c|c|}
\hline
\textbf{Raw Text} & \multicolumn{7}{l|}{\texttt{def calculate\_sum(a, b):}} \\ \hline
\textbf{Tokens} & \texttt{"def"} & \texttt{" calculate"} & \texttt{"\_sum"} & \texttt{"("} & \texttt{"a"} & \texttt{","} & \texttt{" b"} \\ \hline
\textbf{Token IDs} & 8942 & 2651 & 7326 & 91 & 382 & 11 & 404 \\ \hline
\end{tabular}
\caption{Tokenization process: The raw text is split into tokens, which are then mapped to unique integer IDs.}
\label{tab:tokenization_process}
\end{table}

The purpose of tokenization is twofold. First, the number of embeddings required to represent a sequence is reduced. By making each ``atom'' of text contain more characters on average (e.g., a whole word like ``calculate'' instead of individual letters), the computational cost of the model is reduced. The trade-off is that adding more unique tokens increases the vocabulary size and the number of trainable parameters since each unique token needs its own embedding.

Second, concatenating characters into words or subwords allows the resulting embedding to hold more semantic information. While a single character like ``c'' has ambiguous meaning, the token ``calculate'' has a specific semantic value. Once the text is converted into Token IDs, these IDs are used to look up their corresponding vectors in the embedding matrix. 

\newpage

\subsubsection{Transformers}

\begin{figure}[htbp]
\centering
\begin{tikzpicture}[
    font=\small,
    >=stealth,
    % Define a style for the matrices
    matrixnode/.style={
        draw, 
        rectangle, 
        thick,
        fill=blue!5,
        align=center,
        inner sep=0pt
    },
    % Input matrix style (L x d)
    mat_input/.style={matrixnode, minimum height=2.4cm, minimum width=1.5cm},
    % Output matrix style (L x d_k)
    mat_output/.style={matrixnode, minimum height=1.8cm, minimum width=1cm},
    % Style for token labels
    token_label/.style={font=\footnotesize\ttfamily, anchor=east, align=right}
]

    % --- 1. Input Matrix X ---
    \node[mat_input] (X) {};
    \node[above=0.1cm] at (X.north) {$\mathbf{X}$ \scriptsize{(Input)}};
    \node[below=0.1cm] at (X.south) {$L \times d$};

    % Draw horizontal lines inside X to represent tokens
    % assume 4 rows for visualization
    \foreach \i in {1,2,3} {
        \draw[black!40] ($(X.north west)!\i/4!(X.south west)$) -- ($(X.north east)!\i/4!(X.south east)$);
    }

    % Add token labels beside X (aligned with the rows)
    % Row 1 center
    \node[token_label] at ($(X.north west)!0.125!(X.south west)$) {"def"};
    % Row 2 center
    \node[token_label] at ($(X.north west)!0.375!(X.south west)$) {"calculate"};
    % Row 3 center
    \node[token_label] at ($(X.north west)!0.625!(X.south west)$) {"\_sum"};
    % Row 4 center (vdots)
    \node[token_label] at ($(X.north west)!0.875!(X.south west)$) {$\vdots$};

    % --- 2. Resulting Matrices (Q, K, V) ---
    
    % Key (K) - Middle
    \node[mat_output, right=5cm of X] (K) {};
    \node[above=0.1cm] at (K.north) {$\mathbf{K}$ \scriptsize{(Key)}};
    \node[right=0.1cm] at (K.east) {$L \times d_k$};
    % Draw lines inside K
    \foreach \i in {1,2,3} {
        \draw[black!40] ($(K.north west)!\i/4!(K.south west)$) -- ($(K.north east)!\i/4!(K.south east)$);
    }

    % Query (Q) - Above K
    \node[mat_output, above=0.7cm of K] (Q) {};
    \node[above=0.1cm] at (Q.north) {$\mathbf{Q}$ \scriptsize{(Query)}};
    \node[right=0.1cm] at (Q.east) {$L \times d_k$};
    % Draw lines inside Q
    \foreach \i in {1,2,3} {
        \draw[black!40] ($(Q.north west)!\i/4!(Q.south west)$) -- ($(Q.north east)!\i/4!(Q.south east)$);
    }

    % Value (V) - Below K
    \node[mat_output, below=0.7cm of K] (V) {};
    \node[above=0.1cm] at (V.north) {$\mathbf{V}$ \scriptsize{(Value)}};
    \node[right=0.1cm] at (V.east) {$L \times d_v$};
    % Draw lines inside V
    \foreach \i in {1,2,3} {
        \draw[black!40] ($(V.north west)!\i/4!(V.south west)$) -- ($(V.north east)!\i/4!(V.south east)$);
    }

    % --- 3. Arrows and Operations ---
    \coordinate (split) at ($(X.east)!0.5!(K.west)$);

    % Main line out from X
    \draw[thick] (X.east) -- (split);

    % Branch to Q
    \draw[->] (split) |- (Q.west) node[pos=0.75, above] {$\times \mathbf{W}_Q$};
    
    % Branch to K
    \draw[->] (split) -- (K.west) node[midway, above] {$\times \mathbf{W}_K$};
    
    % Branch to V
    \draw[->] (split) |- (V.west) node[pos=0.75, below] {$\times \mathbf{W}_V$};

\end{tikzpicture}
\caption{The Linear Projection step in a Transformer block. The input hidden states $\mathbf{X}$ are multiplied by learned weight matrices to produce the Query, Key, and Value matrices. Note that The matrices shown is input/output matrices, and $\mathbf{W}$ are the matrices that are model-parameters.}
\label{fig:qkv_projection_lines}
\end{figure}

The core component of a large language model is the transformer. A transformer block is a specific neural network architecture; stacked together in sequence (with some operations in parallel inside each block), it forms a deep neural network. The depth can vary in how many layers of transformer blocks it has. Layers (or blocks) are indexed by a superscript. Thus, at layer $\ell$ the hidden states are written as
\[
\mathbf{X}^{(\ell)} \in \mathbb{R}^{L \times d},
\]
and a single transformer block can be viewed as a differentiable function
\[
\mathrm{Transformer}_{\theta}^{(\ell)} : \mathbb{R}^{L \times d} \to \mathbb{R}^{L \times d},
\qquad
\mathbf{X}^{(\ell+1)} = \mathrm{Transformer}_{\theta}^{(\ell)}\!\bigl(\mathbf{X}^{(\ell)}\bigr).
\]
The core strength of the transformer is that it can naturally handle variable-length input and, in a consistent way, learn dependencies between tokens. This was historically difficult in deep learning, since a conventional feed-forward neural network typically assumes a fixed input dimension.

The first step in a self-attention block is to define three parameter matrices
\[
\mathbf{W}_Q,\ \mathbf{W}_K,\ \mathbf{W}_V,
\]
which are shown in Figure~\ref{fig:qkv_projection_lines}. Multiplying these with the input to the model (the token embeddings) gives the query, key and value matrices:
\[
\mathbf{Q} = \mathbf{X}^{(\ell)} \mathbf{W}_Q,\qquad
\mathbf{K} = \mathbf{X}^{(\ell)} \mathbf{W}_K,\qquad
\mathbf{V} = \mathbf{X}^{(\ell)} \mathbf{W}_V,
\]
with
\[
\mathbf{Q}, \mathbf{K} \in \mathbb{R}^{L \times d_k}, 
\qquad
\mathbf{V} \in \mathbb{R}^{L \times d_v}.
\]
Here $L$ is the sequence length and $d_k, d_v$ are the key and value dimensions, respectively. The names "query" and "key" come from database and information-retrieval terminology: in this context they describe how each token "queries" other tokens in the sequence.

\begin{figure}[htbp]
\centering
% move figure left so the wide layout looks centered
\hspace*{-0.19\textwidth}%
\begin{tikzpicture}[
    font=\scriptsize,
    >=stealth,
    node distance=8mm and 8mm,
    boxmatrix/.style={
        draw,
        rectangle,
        thick,
        fill=blue!5,
        align=center,
        inner sep=0pt
    },
    token_label/.style={font=\footnotesize\ttfamily, anchor=east, align=right},
    col_label/.style={font=\footnotesize\ttfamily, anchor=north, rotate=60}
]

% =========================================================
% Q : box with horizontal grid + row labels
% =========================================================
\node[boxmatrix, minimum width=1.0cm, minimum height=3cm] (Q) {$\mathbf{Q}$};

% 7 horizontal lines
\foreach \i in {1,2,3,4,5,6}{
    \draw[black!40]
        ($(Q.north west)!\i/7!(Q.south west)$) --
        ($(Q.north east)!\i/7!(Q.south east)$);
}

% row labels beside Q
\node[token_label] at ($(Q.north west)!0.071!(Q.south west)$) {"def"};
\node[token_label] at ($(Q.north west)!0.214!(Q.south west)$) {"calculate"};
\node[token_label] at ($(Q.north west)!0.357!(Q.south west)$) {"\_sum"};
\node[token_label] at ($(Q.north west)!0.500!(Q.south west)$) {"(a "};
\node[token_label] at ($(Q.north west)!0.643!(Q.south west)$) {", "};
\node[token_label] at ($(Q.north west)!0.786!(Q.south west)$) {"b)"};
\node[token_label] at ($(Q.north west)!0.929!(Q.south west)$) {":\textbackslash n"};

% =========================================================
% ×  K^T : box with vertical grid + column labels
% =========================================================
\node[right=1mm of Q] (times) {$\times$};

\node[boxmatrix, right=1mm of times,
      minimum width=3cm, minimum height=1cm] (KT) {$\mathbf{K}^T$};

% 7 vertical lines
\foreach \i in {1,2,3,4,5,6}{
    \draw[black!40]
        ($(KT.north west)!\i/7!(KT.north east)$) --
        ($(KT.south west)!\i/7!(KT.south east)$);
}

% K^T column labels under grid columns
\node[col_label] at ($(KT.south west)!0.071!(KT.south east)+(-3mm,-4mm)$) {"def"};
\node[col_label] at ($(KT.south west)!0.214!(KT.south east)+(-5mm,-8mm)$) {"calculate"};
\node[col_label] at ($(KT.south west)!0.357!(KT.south east)+(-3mm,-4mm)$) {"\_sum"};
\node[col_label] at ($(KT.south west)!0.500!(KT.south east)+(-3mm,-3.5mm)$) {"(a "};
\node[col_label] at ($(KT.south west)!0.643!(KT.south east)+(-3mm,-3mm)$) {", "};
\node[col_label] at ($(KT.south west)!0.786!(KT.south east)+(-3mm,-3mm)$) {"b)"};
\node[col_label] at ($(KT.south west)!0.929!(KT.south east)+(-3mm,-3mm)$) {":\textbackslash n"};

% =========================================================
% Raw scores box with 7x7 grid + numbers
% =========================================================
\node[right=1mm of KT] (arrow1) {$\xrightarrow{\div\sqrt{d_k}}$};

\node[boxmatrix, minimum width=4.2cm, minimum height=4.2cm, right=10mm of arrow1] (RawBox) {};

% grid in RawBox
\foreach \i in {1,2,3,4,5,6}{
    % horizontal
    \draw[black!40]
      ($(RawBox.north west)!\i/7!(RawBox.south west)$) --
      ($(RawBox.north east)!\i/7!(RawBox.south east)$);
    % vertical
    \draw[black!40]
      ($(RawBox.north west)!\i/7!(RawBox.north east)$) --
      ($(RawBox.south west)!\i/7!(RawBox.south east)$);
}

\node[above=2mm of RawBox.north] {Raw Scores};

% row/column centres for Raw
\foreach \i in {1,...,7}{
    \coordinate (RawRow\i) at ($(RawBox.north west)!{(2*\i-1)/14}!(RawBox.south west)$);
}
\foreach \j in {1,...,7}{
    \coordinate (RawCol\j) at ($(RawBox.north west)!{(2*\j-1)/14}!(RawBox.north east)$);
}

% numbers in Raw (scores)
% row 1 "def"
\node at (RawRow1 -| RawCol1) {4.0};
\node at (RawRow1 -| RawCol2) {1.2};
\node at (RawRow1 -| RawCol3) {0.5};
\node at (RawRow1 -| RawCol4) {-0.3};
\node at (RawRow1 -| RawCol5) {0.8};
\node at (RawRow1 -| RawCol6) {0.2};
\node at (RawRow1 -| RawCol7) {-0.5};
% row 2 "calculate"
\node at (RawRow2 -| RawCol1) {2.5};
\node at (RawRow2 -| RawCol2) {5.0};
\node at (RawRow2 -| RawCol3) {1.8};
\node at (RawRow2 -| RawCol4) {0.3};
\node at (RawRow2 -| RawCol5) {-0.2};
\node at (RawRow2 -| RawCol6) {0.5};
\node at (RawRow2 -| RawCol7) {0.1};
% row 3 "_sum"
\node at (RawRow3 -| RawCol1) {1.2};
\node at (RawRow3 -| RawCol2) {4.5};
\node at (RawRow3 -| RawCol3) {4.8};
\node at (RawRow3 -| RawCol4) {0.8};
\node at (RawRow3 -| RawCol5) {0.2};
\node at (RawRow3 -| RawCol6) {-0.3};
\node at (RawRow3 -| RawCol7) {0.5};
% row 4 "(a "
\node at (RawRow4 -| RawCol1) {0.8};
\node at (RawRow4 -| RawCol2) {3.2};
\node at (RawRow4 -| RawCol3) {2.8};
\node at (RawRow4 -| RawCol4) {3.5};
\node at (RawRow4 -| RawCol5) {0.5};
\node at (RawRow4 -| RawCol6) {0.9};
\node at (RawRow4 -| RawCol7) {0.2};
% row 5 ", "
\node at (RawRow5 -| RawCol1) {0.3};
\node at (RawRow5 -| RawCol2) {1.5};
\node at (RawRow5 -| RawCol3) {1.0};
\node at (RawRow5 -| RawCol4) {3.8};
\node at (RawRow5 -| RawCol5) {2.5};
\node at (RawRow5 -| RawCol6) {0.8};
\node at (RawRow5 -| RawCol7) {0.4};
% row 6 "b)"
\node at (RawRow6 -| RawCol1) {0.5};
\node at (RawRow6 -| RawCol2) {2.0};
\node at (RawRow6 -| RawCol3) {1.5};
\node at (RawRow6 -| RawCol4) {3.0};
\node at (RawRow6 -| RawCol5) {1.8};
\node at (RawRow6 -| RawCol6) {3.2};
\node at (RawRow6 -| RawCol7) {0.6};
% row 7 ":\n"
\node at (RawRow7 -| RawCol1) {1.8};
\node at (RawRow7 -| RawCol2) {2.5};
\node at (RawRow7 -| RawCol3) {2.0};
\node at (RawRow7 -| RawCol4) {1.2};
\node at (RawRow7 -| RawCol5) {0.8};
\node at (RawRow7 -| RawCol6) {1.5};
\node at (RawRow7 -| RawCol7) {2.8};

% row labels for Raw
\node[token_label] at ($(RawRow1)+(0,0)$) {"def"};
\node[token_label] at ($(RawRow2)+(0,0)$) {"calculate"};
\node[token_label] at ($(RawRow3)+(0,0)$) {"\_sum"};
\node[token_label] at ($(RawRow4)+(0,0)$) {"(a "};
\node[token_label] at ($(RawRow5)+(0,0)$) {", "};
\node[token_label] at ($(RawRow6)+(0,0)$) {"b)"};
\node[token_label] at ($(RawRow7)+(0,0)$) {":\textbackslash n"};

% column labels for Raw
\node[col_label] at ($(RawBox.south -| RawCol1)+(-3mm,-4mm)$) {"def"};
\node[col_label] at ($(RawBox.south -| RawCol2)+(-5mm,-8mm)$) {"calculate"};
\node[col_label] at ($(RawBox.south -| RawCol3)+(-3mm,-4mm)$) {"\_sum"};
\node[col_label] at ($(RawBox.south -| RawCol4)+(-3mm,-3.5mm)$) {"(a "};
\node[col_label] at ($(RawBox.south -| RawCol5)+(-3mm,-3mm)$) {", "};
\node[col_label] at ($(RawBox.south -| RawCol6)+(-3mm,-3mm)$) {"b)"};
\node[col_label] at ($(RawBox.south -| RawCol7)+(-3mm,-3mm)$) {":\textbackslash n"};

% =========================================================
% Mask + Softmax arrow
% =========================================================
\draw[->]
    ([xshift=3mm]RawBox.east) -- ++(12mm,0)
    node[midway, above, align=center, font=\scriptsize] {Mask + \\ Softmax}
    coordinate (softarrowend);

% =========================================================
% Attention weights box with 7x7 grid + numbers
% =========================================================
\node[boxmatrix, minimum width=4.2cm, minimum height=4.2cm, right=14mm of softarrowend] (SoftBox) {};

% grid in SoftBox
\foreach \i in {1,2,3,4,5,6}{
    % horizontal
    \draw[black!40]
      ($(SoftBox.north west)!\i/7!(SoftBox.south west)$) --
      ($(SoftBox.north east)!\i/7!(SoftBox.south east)$);
    % vertical
    \draw[black!40]
      ($(SoftBox.north west)!\i/7!(SoftBox.north east)$) --
      ($(SoftBox.south west)!\i/7!(SoftBox.south east)$);
}

\node[above=2mm of SoftBox.north] {Attention Weights};

% row/column centres for SoftBox
\foreach \i in {1,...,7}{
    \coordinate (SoftRow\i) at ($(SoftBox.north west)!{(2*\i-1)/14}!(SoftBox.south west)$);
}
\foreach \j in {1,...,7}{
    \coordinate (SoftCol\j) at ($(SoftBox.north west)!{(2*\j-1)/14}!(SoftBox.north east)$);
}

% numbers in SoftBox (Mask + Softmax of Raw scores, causal mask)
% row 1 "def": softmax([4.0]) = [1.00]
\node at (SoftRow1 -| SoftCol1) {1.00};
\node at (SoftRow1 -| SoftCol2) {0};
\node at (SoftRow1 -| SoftCol3) {0};
\node at (SoftRow1 -| SoftCol4) {0};
\node at (SoftRow1 -| SoftCol5) {0};
\node at (SoftRow1 -| SoftCol6) {0};
\node at (SoftRow1 -| SoftCol7) {0};
% row 2 "calculate": softmax([2.5, 5.0])
\node at (SoftRow2 -| SoftCol1) {.08};
\node at (SoftRow2 -| SoftCol2) {.92};
\node at (SoftRow2 -| SoftCol3) {0};
\node at (SoftRow2 -| SoftCol4) {0};
\node at (SoftRow2 -| SoftCol5) {0};
\node at (SoftRow2 -| SoftCol6) {0};
\node at (SoftRow2 -| SoftCol7) {0};
% row 3 "_sum": softmax([1.2, 4.5, 4.8])
\node at (SoftRow3 -| SoftCol1) {.02};
\node at (SoftRow3 -| SoftCol2) {.42};
\node at (SoftRow3 -| SoftCol3) {.57};
\node at (SoftRow3 -| SoftCol4) {0};
\node at (SoftRow3 -| SoftCol5) {0};
\node at (SoftRow3 -| SoftCol6) {0};
\node at (SoftRow3 -| SoftCol7) {0};
% row 4 "(a ": softmax([0.8, 3.2, 2.8, 3.5])
\node at (SoftRow4 -| SoftCol1) {.03};
\node at (SoftRow4 -| SoftCol2) {.32};
\node at (SoftRow4 -| SoftCol3) {.22};
\node at (SoftRow4 -| SoftCol4) {.43};
\node at (SoftRow4 -| SoftCol5) {0};
\node at (SoftRow4 -| SoftCol6) {0};
\node at (SoftRow4 -| SoftCol7) {0};
% row 5 ", ": softmax([0.3, 1.5, 1.0, 3.8, 2.5])
\node at (SoftRow5 -| SoftCol1) {.02};
\node at (SoftRow5 -| SoftCol2) {.07};
\node at (SoftRow5 -| SoftCol3) {.04};
\node at (SoftRow5 -| SoftCol4) {.68};
\node at (SoftRow5 -| SoftCol5) {.19};
\node at (SoftRow5 -| SoftCol6) {0};
\node at (SoftRow5 -| SoftCol7) {0};
% row 6 "b)": softmax([0.5, 2.0, 1.5, 3.0, 1.8, 3.2])
\node at (SoftRow6 -| SoftCol1) {.03};
\node at (SoftRow6 -| SoftCol2) {.12};
\node at (SoftRow6 -| SoftCol3) {.07};
\node at (SoftRow6 -| SoftCol4) {.31};
\node at (SoftRow6 -| SoftCol5) {.09};
\node at (SoftRow6 -| SoftCol6) {.38};
\node at (SoftRow6 -| SoftCol7) {0};
% row 7 ":\n": softmax([1.8, 2.5, 2.0, 1.2, 0.8, 1.5, 2.8])
\node at (SoftRow7 -| SoftCol1) {.12};
\node at (SoftRow7 -| SoftCol2) {.23};
\node at (SoftRow7 -| SoftCol3) {.14};
\node at (SoftRow7 -| SoftCol4) {.06};
\node at (SoftRow7 -| SoftCol5) {.04};
\node at (SoftRow7 -| SoftCol6) {.09};
\node at (SoftRow7 -| SoftCol7) {.32};

% row labels for Soft
\node[token_label] at ($(SoftRow1)+(0,0)$) {"def"};
\node[token_label] at ($(SoftRow2)+(0,0)$) {"calculate"};
\node[token_label] at ($(SoftRow3)+(0,0)$) {"\_sum"};
\node[token_label] at ($(SoftRow4)+(0,0)$) {"(a "};
\node[token_label] at ($(SoftRow5)+(0,0)$) {", "};
\node[token_label] at ($(SoftRow6)+(0,0)$) {"b)"};
\node[token_label] at ($(SoftRow7)+(0,0)$) {":\textbackslash n"};

% column labels for Soft
\node[col_label] at ($(SoftBox.south -| SoftCol1)+(-3mm,-4mm)$) {"def"};
\node[col_label] at ($(SoftBox.south -| SoftCol2)+(-5mm,-8mm)$) {"calculate"};
\node[col_label] at ($(SoftBox.south -| SoftCol3)+(-3mm,-4mm)$) {"\_sum"};
\node[col_label] at ($(SoftBox.south -| SoftCol4)+(-3mm,-3.5mm)$) {"(a "};
\node[col_label] at ($(SoftBox.south -| SoftCol5)+(-3mm,-3mm)$) {", "};
\node[col_label] at ($(SoftBox.south -| SoftCol6)+(-3mm,-3mm)$) {"b)"};
\node[col_label] at ($(SoftBox.south -| SoftCol7)+(-3mm,-3mm)$) {":\textbackslash n"};

\end{tikzpicture}
\caption{Visualization of how multiplying Q and K generates a score matrix, which encodes how related two tokens are to each other. The mask and softmax normalization normalizes values and preserves causality (autregression traiing objective).}
\label{fig:attention_mechanism_wide}
\end{figure}

By multiplying queries and keys, the (unnormalised) attention scores are obtained,
\[
\mathbf{S} = \frac{1}{\sqrt{d_k}}\,\mathbf{Q}\mathbf{K}^\top \in \mathbb{R}^{L \times L}.
\]
The score matrix $\mathbf{S}$ tells us how all tokens relate to each other: larger positive values indicate that two tokens are more relevant to one another in the current representation space. However, at inference time a language model cannot see future tokens. To enforce this during training as well, a causal mask is applied that hides all "future" positions: for each position $i$, all entries $(i,j)$ with $j > i$ are set to $-\infty$ (and unmasked positions are left unchanged). Let $\mathbf{M} \in \mathbb{R}^{L \times L}$ denote this mask matrix.

After adding the mask, a row-wise softmax is performed. For each row $i$ this is defined by
\[
A_{ij}
=
\frac{\exp\bigl(S_{ij} + M_{ij}\bigr)}{\sum_{k=1}^{L} \exp\bigl(S_{ik} + M_{ik}\bigr)}.
\]
Softmax normalises each row so that the entries sum to exactly $1$, and any entry with $S_{ij} + M_{ij} = -\infty$ is mapped to $0$. In addition to this normalisation, the exponential introduces a non-linearity into the model.

Thus, after masking and a row-wise softmax, the attention matrix is obtained
\[
\mathbf{A} = \operatorname{Softmax}\bigl(\mathbf{S} + \mathbf{M}\bigr) \in \mathbb{R}^{L \times L},
\]
as illustrated in Figure~\ref{fig:attention_mechanism_wide}. Each row of $\mathbf{A}$ encodes how strongly a given token attends to every other token in the sequence, conditioned on the current context.

The attention matrix is multiplied with the value matrix to produce the output
\[
\mathbf{Z} = \mathbf{A}\mathbf{V} \in \mathbb{R}^{L \times d_v},
\]
as seen in Figure~\ref{fig:av_mlp_block}. The resulting matrix $\mathbf{Z}$ implicitly contains information about how each token (along the $L$ dimension) relates to the previous ones, and in which way.

Finally, a small neural network (a multi-layer perceptron) is applied to $\mathbf{Z}$ to add a further learnable, non-linear transformation of this output.

\begin{figure}[htbp]
\centering
\hspace*{-0.1\textwidth}%
\begin{tikzpicture}[
    font=\small,
    >=stealth,
    % Base style for rectangular matrices
    boxmatrix/.style={
        draw,
        rectangle,
        thick,
        fill=blue!5,
        align=center,
        inner sep=0pt
    },
    % Token labels on the side (rows)
    token_label/.style={font=\footnotesize\ttfamily, anchor=east, align=right}
]

% =========================================================
% 1. Attention matrix A (L x L), square, 7x7 grid
% =========================================================
\node[boxmatrix, minimum width=3cm, minimum height=3cm] (A) {$\mathbf{A}$};

% 7x7 grid in A
\foreach \i in {1,2,3,4,5,6}{
    % horizontal
    \draw[black!40]
        ($(A.north west)!\i/7!(A.south west)$) --
        ($(A.north east)!\i/7!(A.south east)$);
    % vertical
    \draw[black!40]
        ($(A.north west)!\i/7!(A.north east)$) --
        ($(A.south west)!\i/7!(A.south east)$);
}

\node[below=0.1cm] at (A.south) {$L \times L$};

% =========================================================
% 2. Value matrix V (L x d), tall and narrow, 7 rows with words
% =========================================================
\node[right=1mm of A] (timesAV) {$\times$};

\node[boxmatrix, right=15mm of timesAV,
      minimum width=1.0cm, minimum height=3cm] (V) {$\mathbf{V}$};

% 7 horizontal lines in V
\foreach \i in {1,2,3,4,5,6}{
    \draw[black!40]
        ($(V.north west)!\i/7!(V.south west)$) --
        ($(V.north east)!\i/7!(V.south east)$);
}

% row labels beside V (7 tokens)
\node[token_label] at ($(V.north west)!0.071!(V.south west)$) {"def"};
\node[token_label] at ($(V.north west)!0.214!(V.south west)$) {"calculate"};
\node[token_label] at ($(V.north west)!0.357!(V.south west)$) {"\_sum"};
\node[token_label] at ($(V.north west)!0.500!(V.south west)$) {"(a "};
\node[token_label] at ($(V.north west)!0.643!(V.south west)$) {", "};
\node[token_label] at ($(V.north west)!0.786!(V.south west)$) {"b)"};
\node[token_label] at ($(V.north west)!0.929!(V.south west)$) {":\textbackslash n"};

\node[below=0.1cm] at (V.south) {$L \times d_{\text{model}}$};

% =========================================================
% 3. Z = A V  (L x d_model)
% =========================================================
\draw[->]
    ([xshift=2mm]V.east) -- ++(1.2cm,0)
    node[midway, above, font=\scriptsize] {$\mathbf{Z} = \mathbf{A}\mathbf{V}$}
    coordinate (Zin);

\node[boxmatrix, right=0.2cm of Zin,
      minimum width=1.0cm, minimum height=3cm] (Z) {$\mathbf{Z}$};

% 7 horizontal lines in Z
\foreach \i in {1,2,3,4,5,6}{
    \draw[black!40]
        ($(Z.north west)!\i/7!(Z.south west)$) --
        ($(Z.north east)!\i/7!(Z.south east)$);
}

\node[below=0.1cm] at (Z.south) {$L \times d_{\text{model}}$};

% =========================================================
% 4. First MLP layer: H = \phi(Z W^{(1)})  (L x 2 d_model)
% =========================================================
\draw[->]
    ([xshift=5mm]Z.east) -- ++(1.3cm,0)
    node[midway, above, font=\scriptsize, align=center]
        {\textcolor{red}{$W^{(1)}$}\\[1pt]$\in \mathbb{R}^{d_{\text{model}} \times 2d_{\text{model}}}$}
    coordinate (Hin);

\node[boxmatrix, right=0.35cm of Hin,
      minimum width=1.8cm, minimum height=3cm] (H) {};

\node[above=0.1cm] at (H.north) {MLP hidden};
\node[below=0.1cm] at (H.south) {$L \times 2d_{\text{model}}$};

% 7 horizontal lines in H
\foreach \i in {1,2,3,4,5,6}{
    \draw[black!40]
        ($(H.north west)!\i/7!(H.south west)$) --
        ($(H.north east)!\i/7!(H.south east)$);
}

% One vertical split to suggest higher width (2d)
\draw[black!30]
    ($(H.north west)!0.5!(H.north east)$) --
    ($(H.south west)!0.5!(H.south east)$);

% =========================================================
% 5. Second MLP layer: X' = H W^{(2)}  (L x d_model)
% =========================================================
\draw[->]
    ([xshift=5mm]H.east) -- ++(1.3cm,0)
    node[midway, above, font=\scriptsize, align=center]
        {\textcolor{red}{$W^{(2)}$}\\[1pt]$\in \mathbb{R}^{2d_{\text{model}} \times d_{\text{model}}}$}
    coordinate (Xoutin);

\node[boxmatrix, right=0.35cm of Xoutin,
      minimum width=1.0cm, minimum height=3cm] (Xout) {$\mathbf{X}'$};

% 7 horizontal lines in X'
\foreach \i in {1,2,3,4,5,6}{
    \draw[black!40]
        ($(Xout.north west)!\i/7!(Xout.south west)$) --
        ($(Xout.north east)!\i/7!(Xout.south east)$);
}

% row labels beside X' (on the RIGHT side now)
\node[token_label, anchor=west, align=left] at ($(Xout.north east)!0.071!(Xout.south east)$) {"def"};
\node[token_label, anchor=west, align=left] at ($(Xout.north east)!0.214!(Xout.south east)$) {"calculate"};
\node[token_label, anchor=west, align=left] at ($(Xout.north east)!0.357!(Xout.south east)$) {"\_sum"};
\node[token_label, anchor=west, align=left] at ($(Xout.north east)!0.500!(Xout.south east)$) {"(a "};
\node[token_label, anchor=west, align=left] at ($(Xout.north east)!0.643!(Xout.south east)$) {", "};
\node[token_label, anchor=west, align=left] at ($(Xout.north east)!0.786!(Xout.south east)$) {"b)"};
\node[token_label, anchor=west, align=left] at ($(Xout.north east)!0.929!(Xout.south east)$) {":\textbackslash n"};

\node[below=0.1cm] at (Xout.south) {$L \times d_{\text{model}}$};

\end{tikzpicture}
\caption{Given the attention matrix, it is multiplied with the value matrix to extract the final contextual representation
$\mathbf{Z}$ from the input. The resulting output is then passed through a neural network (a multi-layer
perceptron), which adds further non-linear and learnable transformations.}
\label{fig:av_mlp_block}
\end{figure}

An important note is that, in practice, multi-head attention is used: instead of a single pair of $(\mathbf{Q}, \mathbf{K}, \mathbf{V})$, several such sets are employed and thus several attention matrices, each of which is multiplied with its corresponding value matrix. This allows each layer to model different types of relationships in parallel. Empirically, different heads are often observed to specialise in attending to different kinds of semantic or positional information~\cite{vaswani2017attention}.

\begin{figure}[htbp]
\centering
\hspace*{-0.1\textwidth}
\begin{tikzpicture}[
    font=\small,
    >=stealth,
    boxmatrix/.style={
        draw,
        rectangle,
        thick,
        fill=blue!5,
        align=center,
        inner sep=0pt
    }
]

% =========================================================
% 1. Embeddings X^(0)
% =========================================================
\node[boxmatrix, minimum width=1.0cm, minimum height=2.0cm] (X0) {$\mathbf{X}^{(0)}$};
\node[above=0.1cm] at (X0.north) {Embeddings};
\node[below=0.1cm] at (X0.south) {$L \times d_{\text{model}}$};

% =========================================================
% 2. First transformer layer: X^(1)
% =========================================================
\node[boxmatrix, minimum width=1.0cm, minimum height=2.0cm, right=1.7cm of X0] (X1) {$\mathbf{X}^{(1)}$};
\node[below=0.1cm] at (X1.south) {$L \times d_{\text{model}}$};

\draw[->]
    (X0.east) -- (X1.west)
    node[midway, above, font=\scriptsize] {Transformer};

% =========================================================
% 3. Second transformer layer: X^(2)
% =========================================================
\node[boxmatrix, minimum width=1.0cm, minimum height=2.0cm, right=1.7cm of X1] (X2) {$\mathbf{X}^{(2)}$};
\node[below=0.1cm] at (X2.south) {$L \times d_{\text{model}}$};

\draw[->]
    (X1.east) -- (X2.west)
    node[midway, above, font=\scriptsize] {Transformer};

% =========================================================
% 4. Ellipsis and final layer X^(N)
% =========================================================
\node[right=1.7cm of X2] (Dots) {$\cdots$};

\node[boxmatrix, minimum width=1.2cm, minimum height=2.0cm, right=0.5cm of Dots] (XN) {$\mathbf{X}^{(N)}$};
\node[below=0.1cm] at (XN.south) {$L \times d_{\text{model}}$};

\draw[->]
    (X2.east) -- (Dots.west)
    node[midway, above, font=\scriptsize] {Transformer};

\draw[->]
    (Dots.east) -- (XN.west)
    node[midway, above, font=\scriptsize] {};

% =========================================================
% 5. Head: X^N -> logits
% =========================================================
\draw[->]
    ([xshift=2mm]XN.east) -- ++(1.5cm,0)
    node[midway, above, font=\scriptsize, align=center]
        {\textcolor{red}{$\times \mathbf{W}_{\text{head}}$}\\[1pt]$\in \mathbb{R}^{d_{\text{model}} \times |V|}$}
    coordinate (LogitIn);

\node[boxmatrix, minimum width=1.4cm, minimum height=2.0cm, right=0.2cm of LogitIn] (Logits) {Logits};
\node[below=0.1cm] at (Logits.south) {$L \times |V|$};

% =========================================================
% 6. Softmax: logits -> probabilities
% =========================================================
\draw[->]
    ([xshift=2mm]Logits.east) -- ++(1.5cm,0)
    node[midway, above, font=\scriptsize] {Softmax}
    coordinate (ProbIn);place

\node[boxmatrix, minimum width=1.2cm, minimum height=2.0cm, right=0.2cm of ProbIn] (Probs) {Probs};
\node[below=0.1cm] at (Probs.south) {$L \times |V|$};

\end{tikzpicture}
\caption{Full pipeline of a LLM. The input is the input embeddings, and as seen in Figure \ref{fig:av_mlp_block}, the output of a transformer block is of same dimension as output. For each layer, the output is added to the input such that the values of the states after each layer changes. The output of the last transformer layer is called the final hidden states. From which a single linear layer produces our predictions for next token.}
\label{fig:full_transformer_pipeline}
\end{figure}

\newpage

\subsubsection{Head}

Going into the final part of the pipeline shown in Figure~\ref{fig:full_transformer_pipeline}, there are a few important observations about how this stage operates, as it is central to this thesis. First, it is noted that the dimension of the final hidden states is proportional to the context length $L$ (the number of tokens in the sequence). In particular, the output of the last transformer layer can be written as
\[
\mathbf{X}^{(N)} \in \mathbb{R}^{L \times d_{\text{model}}},
\]
so that each row $\mathbf{x}^{(N)}_i \in \mathbb{R}^{d_{\text{model}}}$ represents the final hidden state associated with the $i$-th token in the input.

\begin{figure}[htbp]
\centering
\hspace*{-0.13\textwidth}%
\begin{tikzpicture}[
    font=\small,
    >=stealth,
    boxmatrix/.style={
        draw,
        rectangle,
        thick,
        fill=blue!5,
        align=center,
        inner sep=0pt
    },
    token_label/.style={font=\scriptsize\ttfamily, anchor=east, align=right},
    num_style/.style={font=\scriptsize, text=black!80, anchor=center}
]

% =========================================================
% 1. Final Hidden States X^(N)
% =========================================================
\node[boxmatrix, minimum width=0.9cm, minimum height=3.5cm] (XN) {$\mathbf{X}^{(N)}$};
\node[above=0.1cm] at (XN.north) {\scriptsize Hidden states};
\node[below=0.1cm] at (XN.south) {\scriptsize $L \times d$};

% 7 rows
\foreach \i in {1,2,3,4,5,6}{
    \draw[black!40]
        ($(XN.north west)!\i/7!(XN.south west)$) --
        ($(XN.north east)!\i/7!(XN.south east)$);
}

% Input token labels
\node[token_label] at ($(XN.north west)!0.071!(XN.south west)$) {"def"};
\node[token_label] at ($(XN.north west)!0.214!(XN.south west)$) {"calculate"};
\node[token_label] at ($(XN.north west)!0.357!(XN.south west)$) {"\_sum"};
\node[token_label] at ($(XN.north west)!0.500!(XN.south west)$) {"(a "};
\node[token_label] at ($(XN.north west)!0.643!(XN.south west)$) {", "};
\node[token_label] at ($(XN.north west)!0.786!(XN.south west)$) {"b)"};
\node[token_label] at ($(XN.north west)!0.929!(XN.south west)$) {":\textbackslash n"};

% =========================================================
% 2. Head Matrix W_H
% =========================================================
\node[right=0.15cm of XN] (times) {$\times$};

\node[boxmatrix, right=0.15cm of times,
      minimum width=1.0cm, minimum height=1.2cm] (Head) {$\mathbf{W}_{H}$};
\node[above=0.1cm] at (Head.north) {\scriptsize Head};

% =========================================================
% 3. Logits Matrix (7 x 7, with middle "..." column)
% =========================================================
\draw[->]
    (Head.east) -- ++(0.6cm,0)
    coordinate (LogitIn);

\node[boxmatrix, right=0.15cm of LogitIn,
      minimum width=4.6cm, minimum height=3.5cm] (Logits) {};
\node[above=0.1cm] at (Logits.north) {\scriptsize Logits};
\node[below=0.2cm] at (Logits.south) {\scriptsize $L \times |V|$};

% 7 x 7 grid
\foreach \i in {1,2,3,4,5,6}{
    % horizontal
    \draw[black!40]
        ($(Logits.north west)!\i/7!(Logits.south west)$) --
        ($(Logits.north east)!\i/7!(Logits.south east)$);
}
\foreach \j in {1,2,3,4,5,6}{
    % vertical
    \draw[black!40]
        ($(Logits.north west)!\j/7!(Logits.north east)$) --
        ($(Logits.south west)!\j/7!(Logits.south east)$);
}

% Row / column centres for logits
\foreach \i in {1,...,7}{
    \coordinate (Lrow\i) at ($(Logits.north west)!{(2*\i-1)/14}!(Logits.south west)$);
}
\foreach \j in {1,...,7}{
    \coordinate (Lcol\j) at ($(Logits.north west)!{(2*\j-1)/14}!(Logits.north east)$);
}

% Example logits, row-wise:
% Row 1: [1.34, -3.01, 4.21, ..., -0.85, 1.12, 0.47]
\node[num_style] at (Lrow1 -| Lcol1) {$1.34$};
\node[num_style] at (Lrow1 -| Lcol2) {$-3.01$};
\node[num_style] at (Lrow1 -| Lcol3) {$4.21$};
\node[num_style] at (Lrow1 -| Lcol4) {$\cdots$};
\node[num_style] at (Lrow1 -| Lcol5) {$-0.85$};
\node[num_style] at (Lrow1 -| Lcol6) {$1.12$};
\node[num_style] at (Lrow1 -| Lcol7) {$0.47$};

% Row 2: [-0.42, 2.75, 0.09, ..., 1.93, -1.10, 0.38]
\node[num_style] at (Lrow2 -| Lcol1) {$-0.42$};
\node[num_style] at (Lrow2 -| Lcol2) {$2.75$};
\node[num_style] at (Lrow2 -| Lcol3) {$0.09$};
\node[num_style] at (Lrow2 -| Lcol4) {$\cdots$};
\node[num_style] at (Lrow2 -| Lcol5) {$1.93$};
\node[num_style] at (Lrow2 -| Lcol6) {$-1.10$};
\node[num_style] at (Lrow2 -| Lcol7) {$0.38$};

% Row 3: [0.31, -0.87, 3.02, ..., -2.15, 0.64, 1.58]
\node[num_style] at (Lrow3 -| Lcol1) {$0.31$};
\node[num_style] at (Lrow3 -| Lcol2) {$-0.87$};
\node[num_style] at (Lrow3 -| Lcol3) {$3.02$};
\node[num_style] at (Lrow3 -| Lcol4) {$\cdots$};
\node[num_style] at (Lrow3 -| Lcol5) {$-2.15$};
\node[num_style] at (Lrow3 -| Lcol6) {$0.64$};
\node[num_style] at (Lrow3 -| Lcol7) {$1.58$};

% Row 4: [1.01, 0.22, -1.45, ..., 0.77, 2.34, -0.29]
\node[num_style] at (Lrow4 -| Lcol1) {$1.01$};
\node[num_style] at (Lrow4 -| Lcol2) {$0.22$};
\node[num_style] at (Lrow4 -| Lcol3) {$-1.45$};
\node[num_style] at (Lrow4 -| Lcol4) {$\cdots$};
\node[num_style] at (Lrow4 -| Lcol5) {$0.77$};
\node[num_style] at (Lrow4 -| Lcol6) {$2.34$};
\node[num_style] at (Lrow4 -| Lcol7) {$-0.29$};

% Row 5: [-0.73, 1.18, -2.04, ..., 0.94, -0.33, 2.07]
\node[num_style] at (Lrow5 -| Lcol1) {$-0.73$};
\node[num_style] at (Lrow5 -| Lcol2) {$1.18$};
\node[num_style] at (Lrow5 -| Lcol3) {$-2.04$};
\node[num_style] at (Lrow5 -| Lcol4) {$\cdots$};
\node[num_style] at (Lrow5 -| Lcol5) {$0.94$};
\node[num_style] at (Lrow5 -| Lcol6) {$-0.33$};
\node[num_style] at (Lrow5 -| Lcol7) {$2.07$};

% Row 6: [0.55, -1.29, 0.63, ..., 1.41, 0.02, -0.98]
\node[num_style] at (Lrow6 -| Lcol1) {$0.55$};
\node[num_style] at (Lrow6 -| Lcol2) {$-1.29$};
\node[num_style] at (Lrow6 -| Lcol3) {$0.63$};
\node[num_style] at (Lrow6 -| Lcol4) {$\cdots$};
\node[num_style] at (Lrow6 -| Lcol5) {$1.41$};
\node[num_style] at (Lrow6 -| Lcol6) {$0.02$};
\node[num_style] at (Lrow6 -| Lcol7) {$-0.98$};

% Row 7: [1.89, 0.14, -0.57, ..., -1.22, 1.37, 0.05]
\node[num_style] at (Lrow7 -| Lcol1) {$1.89$};
\node[num_style] at (Lrow7 -| Lcol2) {$0.14$};
\node[num_style] at (Lrow7 -| Lcol3) {$-0.57$};
\node[num_style] at (Lrow7 -| Lcol4) {$\cdots$};
\node[num_style] at (Lrow7 -| Lcol5) {$-1.22$};
\node[num_style] at (Lrow7 -| Lcol6) {$1.37$};
\node[num_style] at (Lrow7 -| Lcol7) {$0.05$};

% Column labels (subset of vocab)
\node[font=\scriptsize, anchor=north] at (Lcol1 |- Logits.south) {$v_1$};
\node[font=\scriptsize, anchor=north] at (Lcol2 |- Logits.south) {$v_2$};
\node[font=\scriptsize, anchor=north] at (Lcol3 |- Logits.south) {$v_3$};
\node[font=\scriptsize, anchor=north] at (Lcol4 |- Logits.south) {$\dots$};
\node[font=\scriptsize, anchor=north] at (Lcol5 |- Logits.south) {$v_{|V|-2}$};
\node[font=\scriptsize, anchor=north] at (Lcol6 |- Logits.south) {$v_{|V|-1}$};
\node[font=\scriptsize, anchor=north] at (Lcol7 |- Logits.south) {$v_{|V|}$};

% =========================================================
% 4. Softmax Arrow
% =========================================================
\draw[->]
    (Logits.east) -- ++(1.9cm,0)
    node[midway, above, font=\scriptsize] {Softmax}
    coordinate (ProbIn);

% =========================================================
% 5. Probabilities Matrix (7 x 7)
% =========================================================
\node[boxmatrix, right=0.15cm of ProbIn,
      minimum width=4.6cm, minimum height=3.5cm] (Probs) {};
\node[above=0.1cm] at (Probs.north) {\scriptsize Probabilities};
\node[below=0.2cm] at (Probs.south) {\scriptsize $L \times |V|$};

% 7 x 7 grid
\foreach \i in {1,2,3,4,5,6}{
    % horizontal
    \draw[black!40]
        ($(Probs.north west)!\i/7!(Probs.south west)$) --
        ($(Probs.north east)!\i/7!(Probs.south east)$);
}
\foreach \j in {1,2,3,4,5,6}{
    % vertical
    \draw[black!40]
        ($(Probs.north west)!\j/7!(Probs.north east)$) --
        ($(Probs.south west)!\j/7!(Probs.south east)$);
}

% Row / column centres for probs
\foreach \i in {1,...,7}{
    \coordinate (Prow\i) at ($(Probs.north west)!{(2*\i-1)/14}!(Probs.south west)$);
}
\foreach \j in {1,...,7}{
    \coordinate (Pcol\j) at ($(Probs.north west)!{(2*\j-1)/14}!(Probs.north east)$);
}

% Example probabilities (values < 1; "..." column represents remaining mass)
% Row 1: [0.16, 0.01, 0.42, ..., 0.08, 0.19, 0.14]
\node[num_style] at (Prow1 -| Pcol1) {$0.16$};
\node[num_style] at (Prow1 -| Pcol2) {$0.01$};
\node[num_style] at (Prow1 -| Pcol3) {$0.42$};
\node[num_style] at (Prow1 -| Pcol4) {$\cdots$};
\node[num_style] at (Prow1 -| Pcol5) {$0.08$};
\node[num_style] at (Prow1 -| Pcol6) {$0.19$};
\node[num_style] at (Prow1 -| Pcol7) {$0.14$};

% Row 2: [0.07, 0.55, 0.06, ..., 0.18, 0.05, 0.09]
\node[num_style] at (Prow2 -| Pcol1) {$0.07$};
\node[num_style] at (Prow2 -| Pcol2) {$0.55$};
\node[num_style] at (Prow2 -| Pcol3) {$0.06$};
\node[num_style] at (Prow2 -| Pcol4) {$\cdots$};
\node[num_style] at (Prow2 -| Pcol5) {$0.18$};
\node[num_style] at (Prow2 -| Pcol6) {$0.05$};
\node[num_style] at (Prow2 -| Pcol7) {$0.09$};

% Row 3: [0.04, 0.09, 0.48, ..., 0.06, 0.11, 0.22]
\node[num_style] at (Prow3 -| Pcol1) {$0.04$};
\node[num_style] at (Prow3 -| Pcol2) {$0.09$};
\node[num_style] at (Prow3 -| Pcol3) {$0.48$};
\node[num_style] at (Prow3 -| Pcol4) {$\cdots$};
\node[num_style] at (Prow3 -| Pcol5) {$0.06$};
\node[num_style] at (Prow3 -| Pcol6) {$0.11$};
\node[num_style] at (Prow3 -| Pcol7) {$0.22$};

% Row 4: [0.11, 0.06, 0.03, ..., 0.13, 0.39, 0.28]
\node[num_style] at (Prow4 -| Pcol1) {$0.11$};
\node[num_style] at (Prow4 -| Pcol2) {$0.06$};
\node[num_style] at (Prow4 -| Pcol3) {$0.03$};
\node[num_style] at (Prow4 -| Pcol4) {$\cdots$};
\node[num_style] at (Prow4 -| Pcol5) {$0.13$};
\node[num_style] at (Prow4 -| Pcol6) {$0.39$};
\node[num_style] at (Prow4 -| Pcol7) {$0.28$};

% Row 5: [0.05, 0.21, 0.02, ..., 0.27, 0.09, 0.12]
\node[num_style] at (Prow5 -| Pcol1) {$0.05$};
\node[num_style] at (Prow5 -| Pcol2) {$0.21$};
\node[num_style] at (Prow5 -| Pcol3) {$0.02$};
\node[num_style] at (Prow5 -| Pcol4) {$\cdots$};
\node[num_style] at (Prow5 -| Pcol5) {$0.27$};
\node[num_style] at (Prow5 -| Pcol6) {$0.09$};
\node[num_style] at (Prow5 -| Pcol7) {$0.12$};

% Row 6: [0.10, 0.04, 0.15, ..., 0.24, 0.18, 0.07]
\node[num_style] at (Prow6 -| Pcol1) {$0.10$};
\node[num_style] at (Prow6 -| Pcol2) {$0.04$};
\node[num_style] at (Prow6 -| Pcol3) {$0.15$};
\node[num_style] at (Prow6 -| Pcol4) {$\cdots$};
\node[num_style] at (Prow6 -| Pcol5) {$0.24$};
\node[num_style] at (Prow6 -| Pcol6) {$0.18$};
\node[num_style] at (Prow6 -| Pcol7) {$0.07$};

% Row 7: [0.09, 0.08, 0.05, ..., 0.11, 0.42, 0.13]
\node[num_style] at (Prow7 -| Pcol1) {$0.09$};
\node[num_style] at (Prow7 -| Pcol2) {$0.08$};
\node[num_style] at (Prow7 -| Pcol3) {$0.05$};
\node[num_style] at (Prow7 -| Pcol4) {$\cdots$};
\node[num_style] at (Prow7 -| Pcol5) {$0.11$};
\node[num_style] at (Prow7 -| Pcol6) {$0.42$};
\node[num_style] at (Prow7 -| Pcol7) {$0.13$};

% Column labels for probs (same subset)
\node[font=\scriptsize, anchor=north] at (Pcol1 |- Probs.south) {$v_1$};
\node[font=\scriptsize, anchor=north] at (Pcol2 |- Probs.south) {$v_2$};
\node[font=\scriptsize, anchor=north] at (Pcol3 |- Probs.south) {$v_3$};
\node[font=\scriptsize, anchor=north] at (Pcol4 |- Probs.south) {$\dots$};
\node[font=\scriptsize, anchor=north] at (Pcol5 |- Probs.south) {$v_{|V|-2}$};
\node[font=\scriptsize, anchor=north] at (Pcol6 |- Probs.south) {$v_{|V|-1}$};
\node[font=\scriptsize, anchor=north] at (Pcol7 |- Probs.south) {$v_{|V|}$};

% Output token labels (shifted)
\node[token_label, anchor=west, align=left]
    at ($(Probs.north east)!0.071!(Probs.south east)$) {"calculate"};
\node[token_label, anchor=west, align=left]
    at ($(Probs.north east)!0.214!(Probs.south east)$) {"\_sum"};
\node[token_label, anchor=west, align=left]
    at ($(Probs.north east)!0.357!(Probs.south east)$) {"(a "};
\node[token_label, anchor=west, align=left]
    at ($(Probs.north east)!0.500!(Probs.south east)$) {", "};
\node[token_label, anchor=west, align=left]
    at ($(Probs.north east)!0.643!(Probs.south east)$) {"b)"};
\node[token_label, anchor=west, align=left]
    at ($(Probs.north east)!0.786!(Probs.south east)$) {":\textbackslash n"};
\node[token_label, anchor=west, align=left, text=blue!70!black]
    at ($(Probs.north east)!0.929!(Probs.south east)$) {\emph{next}};

\end{tikzpicture}
\caption{The full projection pipeline. Final hidden states are multiplied by the head matrix to produce logits over the vocabulary. A row-wise Softmax converts these logits into probability distributions over tokens at each position.}
\label{fig:final_projection_full}
\end{figure}

The final part is illustrated in Figure \ref{fig:final_projection_full}. For the output layer, a set of output embeddings is maintained, one for each token in the vocabulary $V$, each living in the same $d_{\text{model}}$-dimensional space. Concretely, this is implemented by the head matrix
\[
\mathbf{W}_{\text{head}} \in \mathbb{R}^{d_{\text{model}} \times |V|},
\]
where the $v$-th column can be viewed as an output embedding $\mathbf{e}_v \in \mathbb{R}^{d_{\text{model}}}$ associated with token $v \in V$. For each hidden state $\mathbf{x}^{(N)}_i$, the dot product is computed with all output embeddings; in matrix form this is the single multiplication
\[
\mathbf{Z} = \mathbf{X}^{(N)} \mathbf{W}_{\text{head}} \in \mathbb{R}^{L \times |V|}.
\]
Thus the logit $Z_{i,v}$ is precisely the dot product between the hidden state at position $i$ and the output embedding of token $v$. Intuitively, the angle and length relationships in this vector space determine which token will be predicted next: tokens whose output embeddings are more aligned with $\mathbf{x}^{(N)}_i$ receive higher logits, and tokens with similar embeddings tend to be assigned similar probabilities. From an engineering perspective, extending the output vocabulary corresponds to adding more columns to $\mathbf{W}_{\text{head}}$ and associating each new column with a new token ID.

The logits are then converted into probabilities via a row-wise Softmax,
\[
q_\theta(t_i = v \mid t_{<i})
=
\frac{\exp(Z_{i,v})}{\sum_{u \in V} \exp(Z_{i,u})},
\]
so that each row of $\mathbf{Z}$ is mapped to a valid probability distribution over the vocabulary. This is analogous to the use of Softmax in the attention mechanism (Figure~\ref{fig:attention_mechanism_wide}), where it normalises the attention scores row-wise. However, the semantic role of Softmax differs in the two cases. In the attention block, Softmax converts similarity scores between tokens into attention weights over the context. In the final output layer, Softmax is what enforces that the model's predictions form a proper probability distribution over next tokens, with the entries summing to $1$ by construction, as in Equation~\ref{eq:model_likelihood}.

It is also worth contrasting training and inference with respect to how these computations are used. The example above corresponds to the training regime: for a sequence of length $L$, a distribution over the vocabulary is produced by the model at \emph{every} position in the sequence, and the training objective (Equation~\ref{eq:avg_token_loss}) aggregates the per-token losses across all positions. This ``parallel prediction'' over the entire sequence allows training to be implemented as large batched matrix operations, which are highly parallelisable on modern GPUs.

At inference time, by contrast, generation is autoregressive. Starting from an initial prompt, $\mathbf{X}^{(N)}$ and the corresponding probability distribution over the vocabulary are computed by the model only for the last position, a single next token is sampled (or selected), and then this token is appended to the input sequence. This updated sequence is embedded again to form a new $\mathbf{X}^{(0)}$ (as in Figure~\ref{fig:full_transformer_pipeline}), and another forward pass produces probabilities for the next token. This process is repeated step by step until an end-of-sequence token is generated or another stopping criterion is met.

\newpage

\subsection{Additional Concepts and Techniques}

Having covered the mathematical foundations of LLMs, deep learning, and the specific architectural construction of a transformer-based language model, the following subsections will introduce more advanced theoretical concepts and practical techniques that this thesis relies on. These include methods and design choices that extend beyond the standard transformer pipeline, and which are particularly relevant for the embedding-focused experiments proposed in this work.

\subsubsection{Kullback--Leibler Divergence}
\label{subsubsec:kl_divergence}

The conventional training objective for language models is the cross-entropy loss, as defined in Equation~\ref{eq:avg_token_loss}. It is well suited to the empirical data distribution $p_{\text{data}}$, where each target is represented as a one-hot vector and all probability mass is placed on the observed token. During training, however, the model learns a full probability distribution over the vocabulary for each context.

Let us now assume a slightly more general setting: there is a known \emph{target} distribution $p$ and a model distribution $q$ that one would like to match to $p$. The cross-entropy between $p$ and $q$ is
\begin{equation*}
    H(p, q)
    = - \sum_{t \in V} p(t) \log q(t).
    \label{eq:ce_p_q}
\end{equation*}
It is possible to rewrite this expression to separate a term that only depends on $p$ from a term that measures the mismatch between $p$ and $q$. To do this, add and subtract $\log p(t)$ inside the sum:
\begin{align*}
    H(p, q)
    &= - \sum_{t \in V} p(t) \log q(t) \\
    &= - \sum_{t \in V} p(t)
       \bigl[\log p(t) + \log \tfrac{q(t)}{p(t)}\bigr] \\
    &= - \sum_{t \in V} p(t) \log p(t)
       \;-\; \sum_{t \in V} p(t) \log \frac{q(t)}{p(t)}.
\end{align*}
The first term is exactly the entropy of $p$,
\begin{equation*}
    H(p) = - \sum_{t \in V} p(t) \log p(t),
\end{equation*}
while the second term can be written as
\begin{equation*}
    - \sum_{t \in V} p(t) \log \frac{q(t)}{p(t)}
    = \sum_{t \in V} p(t) \log \frac{p(t)}{q(t)}.
\end{equation*}
This yields the standard decomposition
\begin{equation*}
    H(p, q)
    = H(p) + D_{\mathrm{KL}}(p \,\|\, q),
\end{equation*}
where the \emph{Kullback--Leibler divergence} between $p$ and $q$ is defined as
\begin{equation}
    D_{\mathrm{KL}}(p \,\|\, q)
    = \sum_{t \in V} p(t) \log \frac{p(t)}{q(t)}.
    \label{eq:kl_def}
\end{equation}

From this expression it can be seen that even if $p = q$, the cross-entropy $H(p, q)$ does not become zero; its minimum value is the entropy $H(p)$. Subtracting $H(p)$ essentially ``renormalises'' the scale so that the minimum value becomes
\[
    D_{\mathrm{KL}}(p \,\|\, q) = 0
    \quad \Longleftrightarrow \quad p = q,
\]
and any deviation of $q$ from $p$ yields a strictly positive value. In other words, KL divergence is a discrepancy measure derived from cross-entropy whose floor is at zero when the two distributions match exactly.

In the LLM setting, $p$ is identified with a target next-token distribution (for example $p_{\text{data}}(\cdot \mid t_{<i})$) and $q$ with the model distribution $q_\theta(\cdot \mid t_{<i})$. Since $H(p)$ does not depend on the model parameters $\theta$, adding or subtracting this term only adds a constant scalar to the loss and therefore does not affect gradients. Minimising the expected cross-entropy and minimising the expected KL divergence induce identical training dynamics:
\begin{align*}
    \arg\min_\theta \,\mathbb{E}\bigl[H(p, q_\theta)\bigr]
    &= \arg\min_\theta \,\mathbb{E}\bigl[H(p) + D_{\mathrm{KL}}(p \,\|\, q_\theta)\bigr] \\
    &= \arg\min_\theta \,\mathbb{E}\bigl[D_{\mathrm{KL}}(p \,\|\, q_\theta)\bigr].
\end{align*}

In the special case used for standard next-token prediction, $p$ is a one-hot vector and $H(p) = 0$. Then the per-token cross-entropy reduces to $-\log q_\theta(t_i \mid t_{<i})$ and coincides numerically with $D_{\mathrm{KL}}(p \,\|\, q_\theta)$. This corresponds to the curves shown in Figure~\ref{fig:loss_gradient}, where only the probability assigned to the single target token appears.

When $p$ is instead a \emph{soft} distribution (for example in distillation or label smoothing), KL divergence makes explicit that the loss now aggregates contributions from all tokens with non-zero probability under $p$. Rather than only receiving a learning signal for the single most likely token, a graded signal is received by the model for every token that the target distribution considers plausible, shaping the full distribution over the vocabulary.

\subsubsection{Knowledge Distillation}
\label{subsubsec:knowledge_distillation}

A method that has become highly relevant with the rise of LLMs is \emph{knowledge distillation}~\cite{hinton2015distilling}. In its standard form, rather than training directly on hard labels from the dataset, a \emph{student} model is trained to reproduce the output distribution of an already trained \emph{teacher} model. In the context of language models, this means that instead of only learning from one-hot next-token targets, the student learns to match the full next-token distribution produced by another LLM.

Let $p_T(t_i \mid t^{(s)}_{<i})$ denote the teacher distribution over the vocabulary $V$ for token position $i$ in sequence $s$, and let $q_\theta(t_i \mid t^{(s)}_{<i})$ be the corresponding student distribution. Reusing the notation from Equation~\ref{eq:avg_token_loss}, a pure \emph{distillation loss} can be defined by replacing the empirical one-hot targets with the teacher distribution in the KL form:
\begin{equation}
    \mathcal{L}_{\mathrm{KD}}(\theta)
    = \frac{1}{M} \sum_{s=1}^N \sum_{i=1}^{L_s}
      D_{\mathrm{KL}}\!\bigl(p_T(\cdot \mid t^{(s)}_{<i}) \,\|\, q_\theta(\cdot \mid t^{(s)}_{<i})\bigr),
    \label{eq:kd_loss}
\end{equation}
where $M = \sum_{s=1}^N L_s$ is the total number of tokens in the corpus and $D_{\mathrm{KL}}$ is given by Equation~\ref{eq:kl_def}. Expanding this expression yields
\begin{equation}
    \mathcal{L}_{\mathrm{KD}}(\theta)
    = \frac{1}{M} \sum_{s=1}^N \sum_{i=1}^{L_s}
      \sum_{t \in V} p_T\bigl(t \mid t^{(s)}_{<i}\bigr)
      \log \frac{p_T\bigl(t \mid t^{(s)}_{<i}\bigr)}
                {q_\theta\bigl(t \mid t^{(s)}_{<i}\bigr)}.
\end{equation}
In words, the student is trained to minimise the KL divergence between its output distribution and that of the teacher, at every position in every training sequence.

In practice, it is common to combine this distillation objective with the original cross-entropy loss on the hard labels. If the standard cross-entropy loss (Equation~\ref{eq:avg_token_loss}) is denoted by $\mathcal{L}_{\mathrm{CE}}(\theta)$, a simple weighted combination is
\begin{equation}
    \mathcal{L}_{\mathrm{total}}(\theta)
    = \lambda \,\mathcal{L}_{\mathrm{CE}}(\theta)
      + (1 - \lambda)\,\mathcal{L}_{\mathrm{KD}}(\theta),
    \qquad \lambda \in [0,1].
    \label{eq:kd_mixed_loss}
\end{equation}
Typically, this is used when training a smaller student model in the presence of a larger teacher, but similar KL-based regularisation can also be applied when fine-tuning a model while constraining it not to deviate too far from a fixed reference distribution (for example, a frozen base model). This pattern is central in RLHF-style training where a policy is penalised for drifting away from a pretrained reference model~\cite{ouyang2022training}, and in recent reasoning-focused systems such as DeepSeek-R1, which anchor a reasoning policy to a base model via KL regularisation in their RL objective~\cite{guo2025deepseekr1}. Empirically, it has been observed in many works that knowledge distillation can yield student models that perform better than if they were trained purely with cross-entropy on hard labels, due to the richer signal provided by the teacher's soft targets~\cite{furlanello2018born}.

A key limitation of conventional knowledge distillation in the LLM setting is that teacher and student are assumed to share the \emph{exact same tokenizer}, and thus the same vocabulary. There are two closely related reasons for this:
\begin{itemize}
    \item From the definition of KL divergence in Equation~\ref{eq:kl_def}, $p_T$ and $q_\theta$ must be distributions over the same support. If the vocabularies differ, the indices refer to different tokens, and comparing $p_T(t)$ and $q_\theta(t)$ becomes meaningless.
    \item Different tokenizers segment the same sentence into different token sequences. This breaks the one-to-one correspondence between positions $i$ in the teacher and student sequences, so there is no straightforward way to define a per-token KL divergence over aligned positions.
\end{itemize}

Consequently, in the traditional formulation, introducing a new tokenizer or modifying the vocabulary prevents direct application of standard distillation losses. As a foreshadowing for the reader, one of the contributions of this thesis is to show that, in certain special cases, distillation-like objectives can nonetheless be constructed that remain well-defined even when the student uses an extended or modified vocabulary.

\subsubsection{Partial Unfreezing}

For a model with a set of parameters $\theta$, the conventional way of training a deep neural network is to calculate the gradient of the loss with respect to \emph{all} parameters and update them jointly. On the other hand, if a pretrained model is started from and fine-tuned on a new task, there are situations where only a subset of the parameters is updated and the others are left unchanged. This is commonly referred to as \emph{partial unfreezing}. Partial unfreezing can be particularly useful when the available fine-tuning data are limited, as fully updating the entire model in such regimes increases the risk of overfitting and catastrophic forgetting of the pretrained capabilities~\cite{howard2018ulmfit}.

No universal rule exists that tells when partial unfreezing should be applied, which modules (tensors) should be unfrozen, or in what order. Much of the practice comes from heuristics and empirical best practices, and also depends on how and where the architecture is changed. If the architecture is changed by adding or modifying modules in the model, it is useful to initialise the new parameters in such a way that the model is not too disturbed in the initial phase of training, for instance by setting the values of the new matrices so that they have only a small impact on the output at the beginning. Secondly, it is generally a good idea to start by unfreezing only the new modules so that they first ``adapt'' to the model in which they are inserted, before unfreezing other parts of the network if needed. Then, as a final step of fine-tuning, additional modules can optionally be unfrozen if the task requires it.

On the other hand, if all modules were to be unfrozen immediately after adding new tensors to the model, the optimisation process would try to adapt to what is essentially random noise that has been injected into the network. This can degrade performance as the model moves away from a good pretrained solution and overfits the small fine-tuning dataset~\cite{howard2018ulmfit}. In short, the process of adding new modules and fine-tuning a foundation model is delicate, and the goal is often to preserve the original capabilities while adding domain-specific behaviour. Another practical advantage of partial unfreezing is that memory usage is reduced compared to fine-tuning the full model, since gradients and optimizer states only have to be stored for a subset of the parameters.

\subsubsection{Low-Rank Adaptation}

Low-Rank Adaptation (LoRA) is a parameter-efficient method for fine-tuning LLMs in particular~\cite{hu2022lora}. In a deep neural network such as an LLM, each component or module with parameters can be viewed as a weight matrix. A difficulty when fine-tuning a large model is that the updates often need to be regularised so that the model does not stray too far from the original pretrained weights. One idea is partial unfreezing, but in many cases \emph{all} modules still need to be modified, since each transformer layer encodes specific knowledge. LoRA addresses this by introducing a low-rank update to the existing weight matrices.

Let $\mathbf{W} \in \mathbb{R}^{d_{\text{out}} \times d_{\text{in}}}$ be a pretrained weight matrix that is to be adapted. In standard fine-tuning, $\mathbf{W}$ itself is updated. In LoRA, $\mathbf{W}$ is kept frozen, and instead two trainable matrices are introduced
\[
    \mathbf{A} \in \mathbb{R}^{d_{\text{out}} \times r},
    \qquad
    \mathbf{B} \in \mathbb{R}^{r \times d_{\text{in}}},
\]
where $r \ll \min(d_{\text{out}}, d_{\text{in}})$ is the \emph{rank} of the adaptation. The effective weight used during the forward pass is
\begin{equation}
    \mathbf{W}' 
    = \mathbf{W} + \Delta \mathbf{W},
    \qquad
    \Delta \mathbf{W} = \mathbf{A}\mathbf{B}.
    \label{eq:lora_update}
\end{equation}
In practice, $\mathbf{A}$ and $\mathbf{B}$ are initialised so that $\Delta \mathbf{W}$ is (approximately) zero at the start of fine-tuning (for example by setting one of the matrices to zero), ensuring that the pretrained behaviour is initially preserved and only gradually modified as training proceeds.

A major advantage of this construction, besides acting as a form of regularisation towards the original weights, is that memory usage during training is greatly reduced. Instead of storing gradients and optimizer states for all $d_{\text{out}} \times d_{\text{in}}$ entries of $\mathbf{W}$, they only need to be stored for the low-rank factors $\mathbf{A}$ and $\mathbf{B}$, which together contain $r(d_{\text{out}} + d_{\text{in}})$ parameters. For typical choices of $r$, this is several orders of magnitude fewer trainable parameters than full fine-tuning, while still allowing every module in the network to be adapted through the low-rank update.

\subsection{Related Works}

The following subsection lists the most relevant works in the field on the topic of vocabulary expansion. The most recent work was published in May 2025, the same month as the conclusion of the experiments in this thesis.

\subsubsection{FOCUS}

One early method proposed for vocabulary expansion is FOCUS (Fast Overlapping Token Combinations Using Sparsemax)~\cite{dobler2023focus}. The key idea is to train a separate embedding model $E_{\text{aux}}$ (typically fastText or a similar static embedding model) on the target-domain data $D$ to obtain semantically meaningful vector representations. The training objective of such embedding models is often based on maximizing the likelihood of a word given its context. For instance, in the case of skip-gram with negative sampling, the objective is
\begin{equation}
    \max_\theta \sum_{(w, c) \in D} \log \sigma(e_w^\top e_c)
    + \sum_{(w, c') \notin D} \log \sigma(-e_w^\top e_{c'}),
\end{equation}
where $(w, c)$ are positive word–context pairs, $(w, c')$ are negative samples, and $e_w, e_c$ denote their embeddings. The practical result of this training is that semantically similar tokens have embeddings with high cosine similarity, while dissimilar ones tend toward negative similarity.

To initialize a new token $t_{\text{new}}$ (e.g.\ \texttt{numpy}), FOCUS represents it as a weighted linear combination of existing source tokens (e.g.\ \texttt{num}, \texttt{py}) in the overlap between the source and target vocabularies:
\begin{equation}
    e_{\text{new}} = w_1 e_{\text{num}} + w_2 e_{\text{py}},
\end{equation}
where $e_{\text{num}}, e_{\text{py}}$ are the \emph{LLM} embeddings of the overlapping tokens, and $w_1, w_2$ are weights derived from similarity scores between \texttt{numpy} and the source tokens in the auxiliary embedding space $E_{\text{aux}}$ (normalised using a sparse transformation such as Sparsemax). In this way, FOCUS aims to place $e_{\text{new}}$ in a direction that reflects the semantics of the most similar overlapping tokens.

While this method offers a semantically grounded initialization, it is limited in two ways: (1) the weights $w_1, w_2$ are computed heuristically from auxiliary similarities rather than being learned jointly with the LLM, and (2) the auxiliary embedding space $E_{\text{aux}}$ is not guaranteed to align with the model’s native embedding space, which can lead to suboptimal initialization, especially in low-resource settings.

\subsubsection{Vocabulary Expansion with 0.01GB}

A more recent work by Yamaguchi et al.~\cite{yamaguchi2024effectivelyexpandvocabularyllms} systematically evaluates various vocabulary initialization strategies in an extremely low-resource setting, using only 30K sentences (approximately 0.01GB of data) from the target language. The study compares initialization methods including \emph{Random}, \emph{Mean}, \emph{FOCUS}, and a novel method called \emph{Align}, across ten typologically diverse languages and two generation tasks (machine translation and summarization).

In the Align method, each sentence $x \in D$ is tokenized using both the source tokenizer $T_{\text{original}}$ and the extended tokenizer $T_{\text{extended}}$. For every new token $t_{\text{new}}$ in the extended vocabulary, its occurrences in $x$ are aligned to the corresponding tokenizations under $T_{\text{original}}$, yielding a collection of token tuples $u = (t_1, t_2, \dots)$ together with their empirical frequencies $f_u$. Let $\mathcal{T}_{t_{\text{new}}}$ denote the set of such aligned tuples. The new embedding is then initialized as a frequency-weighted average over these tuples:
\begin{equation}
    e_{t_{\text{new}}}
    =
    \sum_{u \in \mathcal{T}_{t_{\text{new}}}} 
        f_u \cdot \frac{1}{|u|} \sum_{t' \in u} e_{t'},
\end{equation}
where $e_{t'}$ are the original LLM embeddings of the constituent subtokens $t'$.

Empirically, Yamaguchi et al.\ find that Align and Mean generally outperform FOCUS and Random in terms of both language-model loss and downstream task performance in this low-resource regime, especially when combined with suitable fine-tuning strategies such as LoRA-based adaptation and appropriately chosen (typically shorter) sequence lengths. In summary, the work demonstrates that vocabulary expansion is feasible with very limited target-language data, but that downstream performance depends critically on both the initialization scheme and the subsequent fine-tuning strategy.

\subsubsection{AweDist}

Recently, AweDist~\cite{dobler2025awedist} introduces a novel approach to vocabulary expansion through what the authors call an ``attention-aware'' mechanism, which can be interpreted as keeping track of which tokens relate to each other despite different tokenization schemes.

The core idea is to create token mappings between sequences of text tokenized with and without the expanded tokenizer. For any given text, AweDist generates two tokenization sequences: one using the original tokenizer and another using the expanded tokenizer containing the new tokens. The method then establishes position mappings between these sequences, identifying regions where tokenizations align (same tokens) and diverge (new merged tokens).

\begin{table}[h]
\hspace{-1.5cm}
\begin{tabular}{|l|l|l|}
\hline
\textbf{Tokenization Type} & \textbf{Token Sequence} & \textbf{Position Mapping} \\
\hline
Original & [``The'', ``patient'', ``took'', ``pal'', ``at'', ``able'', ``medication''] & [0, 1, 2, 3, 4, 5, 6] \\
\hline
Expanded & [``The'', ``patient'', ``took'', ``palatable'', ``medication''] & [0, 1, 2, 3, 4] \\
\hline
\multicolumn{3}{|c|}{\textbf{Position Mappings}} \\
\hline
Same tokens & (0,0), (1,1), (2,2), (4,6) & Context preservation \\
\hline
Divergent tokens & (3) $\leftrightarrow$ (3,4,5) & New token vs.\ subtokens \\
\hline
\end{tabular}
\caption{Illustration of token mappings in AweDist. The expanded tokenizer merges ``pal'', ``at'', ``able'' into a single ``palatable'' token, creating divergent mappings while preserving context token alignments.}
\label{tab:awedist_mappings}
\end{table}

For knowledge distillation, AweDist employs a teacher–student framework where the original model serves as the teacher. The method extracts hidden states from both tokenization sequences at a specified layer $\ell$ and optimizes the new token embeddings to minimize the mean squared error between corresponding positions:
\begin{equation}
    L_{\text{AweDist}}
    = \sum_{(i,j) \in M}
      \left\|\mathbf{X}^{(\ell)}_{\text{new}}[i]
             - \mathbf{X}^{(\ell)}_{\text{orig}}[j]\right\|^2,
\end{equation}
where $M$ represents the set of position mappings between the two sequences, $\mathbf{X}^{(\ell)}_{\text{new}}$ are hidden states from the sequence with new tokens, and $\mathbf{X}^{(\ell)}_{\text{orig}}$ are hidden states from the original tokenization. This notation is consistent with the convention in Section~\ref{fig:full_transformer_pipeline}, where $\mathbf{X}^{(\ell)}$ denotes the hidden states at layer $\ell$.

While this method represents a significant advancement in utilizing the original model as a teacher for tokens that remain the same, several limitations and assumptions merit consideration:

\begin{itemize}
    \item \textbf{Only treating input embeddings}: The paper does not address how to train the output head to generate new tokens, asserting this is not strictly necessary. This assumption is problematic in most practical contexts, since differing input/output tokenization creates a mismatch between training and inference objectives, in addition to various efficiency complications for doing inference at scale. When tokenization differs between training and inference phases, the model’s learned probability distributions may not align with the actual token space during generation.
    
    \item \textbf{Heuristic loss for divergent sequences}: While comparing hidden states for identical tokenizations is well motivated, the loss function for divergent token sequences (new token vs.\ subtokens) relies on heuristics that may not precisely align with the goal of producing correct output tokens.
    
    \item \textbf{MSE as proxy loss}: Mean squared error on hidden states serves only as a proxy for the actual objective, which should be correct output generation or, in a teacher–student setup, KL divergence between probability distributions. MSE treats all elements equally and focuses on magnitude rather than direction, which is suboptimal given that attention mechanisms are multiplicative in nature. Methods such as cosine loss, which preserve directional relationships, may be more appropriate since the sign and relative orientation of elements often matter more than their absolute magnitude in attention computations.
\end{itemize}

Despite these limitations, the paper introduces a novel framework for creating divergent and similar token sequences that allows for a student–teacher setup. It addresses one of the key issues with using a student–teacher setup for vocabulary expansion—namely that the context length differs and individual tokens need to be paired between sequences so that they correspond to the same locations in the text. However, one fundamental challenge remains: normally one compares \emph{output} distributions between LLMs and try to match those, but given that the output space is different (due to different numbers of tokens available), it cannot be compared directly. The proposed workaround involves omitting the output space and operating on hidden states as a proxy.

% methods.tex

\section{Methods}

This section describes the methods and experimental setup used in this thesis, as well as the motivations behind the design choices. After reading this section, the reader should understand

\begin{itemize}
    \item which data and model are used,
    \item the proposed method and the core experiment of the thesis, and
    \item the hyperparameters, hardware, and other configuration details required for reproducibility.
\end{itemize}

\subsection{Data and Models}

To conduct an experiment with a large language model, both a model and data to train it on are required. In this thesis, a pretrained model is taken as a starting point and fine-tuned on a limited corpus of data. The choice of model and dataset is motivated in this subsection.

\subsubsection{Data}

While Ericsson uses its own proprietary programming language, the training data used for this experiment is \texttt{CodeFeedback-Filtered-Instruction}, which is a mix of multiple programming languages including Python, C++, Java, and JavaScript~\cite{zheng2024opencodeinterpreter}. From this dataset, a subset of 1000 query–answer pairs was sampled, and the new tokenizer for vocabulary expansion was trained on this subset. Subsequently, a subset of 5000 files was selected as training data for the LLM and another 10000 files as a validation set. The relatively small sample sizes were chosen both to reduce the total amount of data required for training and to adhere to the central constraint of this work: performing vocabulary expansion in a small-data regime, which is the engineering challenge under consideration.

\newpage

Below is an example query–answer pair from the \texttt{CodeFeedback-Filtered-Instruction} dataset:

\textbf{Query:} Write a function that takes a list of numbers and returns a dictionary with the count of each number.

\textbf{Answer:}
\begin{lstlisting}[language=Python]
def count_numbers(numbers):
    count_dict = {}
    for num in numbers:
        if num in count_dict:
            count_dict[num] += 1
        else:
            count_dict[num] = 1
    return count_dict
\end{lstlisting}

It is also worth noting that most query–answer pairs in the dataset are much longer, often in the thousands of tokens rather than the few dozen tokens shown in this illustrative example.

\subsubsection{Model}

The experiments use the DeepSeek Coder--7B-Instruct-v1.5 model, a 7~billion-parameter transformer decoder with a model dimension of 4096 and a vocabulary of 32\,022 tokens (32\,K subwords + 22 formatting tokens). The base DeepSeek Coder--7B model was pretrained from scratch on 2~trillion tokens (87\,\% code, 13\,\% natural language) using a 16\,K-token context window and an auxiliary fill-in-the-blank task. It was then further pretrained on 2~trillion tokens with a 4\,K-token window under a next-token prediction objective, before being fine-tuned on 2~billion tokens of instruction data to yield the Instruct-v1.5 variant~\cite{sha2024deepseek}.

This model was chosen because it achieves strong benchmark performance while remaining small enough to be trained without sharding or massive compute clusters. Furthermore, it does not use tied weights between the input embedding table and the final output head, meaning that the final projection matrix and the embedding matrix are distinct. Weight tying reduces memory usage by reusing the same matrix for both the input embeddings and the final head $\mathbb{R}^{d \times |V|}$, where $d$ is the model dimension and $|V|$ is the vocabulary size. While weight tying is common in older and smaller models and simplifies vocabulary expansion (since only a single module needs to be initialized and trained), it is less representative of many larger, modern LLMs. It is worth noting that much of the existing literature, including~\cite{dobler2023focus,yamaguchi2024effectivelyexpandvocabularyllms}, assumes weight tying. The method proposed in this thesis therefore goes beyond previous work in both scope and complexity by explicitly handling the untied embedding and output head.

\subsubsection{Tokenization}

Each LLM comes with its own tokenizer; in this case, the model is equipped with a vocabulary of 32\,K tokens and 22 formatting tokens. This subsection explains how additional tokens were introduced and the motivation behind the chosen approach.

Given the training data, together with the model and tokenizer provided via Hugging Face, the goal is to extend the existing tokenizer with new tokens. A natural way to add vocabulary would be to continue training the original tokenizer. For example, if the tokens \texttt{"num"} and \texttt{"py"} already exist, further training could introduce a new token \texttt{"numpy"} with a merge rule \([\texttt{"num"}, \texttt{"py"}]\). However, the standard tokenizer utilities in Hugging Face do not support such incremental training seamlessly. In particular, continuing training can change the indices of existing tokens, which is unacceptable because the LLM has been trained under the assumption that each token index corresponds to a fixed embedding and output weight.

For this reason, new tokens are added manually as special tokens. Concretely, a new tokenizer is first trained on the selected subset of the dataset, and tokens that are frequent in this new tokenizer are then added as special tokens to the original tokenizer. While this method is straightforward, it comes with certain caveats, most notably that no new merge rules involving the added tokens are introduced. For example, if a new token \texttt{"stand"} is added while an existing token \texttt{"standard"} is already present, then tokenizing the string \texttt{"standard"} may produce \([\texttt{"stand"}, \texttt{"ard"}]\) rather than \([\texttt{"standard"}]\), since the merge rule \([\texttt{"stand"}, \texttt{"ard"}]\) does not exist. To avoid degrading the effective tokenization quality (and thus density) when expanding the vocabulary, tokens that are substrings of existing tokens are therefore not added. The full procedure is summarized in Algorithm~\ref{alg:vocab-expansion}.

For the experiments conducted in this thesis, an expanded vocabulary of 800 additional tokens was used.

\subsection{Proposed Method}

Given a pretrained model with its original tokenizer, together with an extended tokenizer containing additional tokens, this subsection outlines the proposed method for vocabulary expansion of an LLM with non-tied model weights. The method specifies how the new input embeddings and the corresponding rows in the final output head are initialized, and motivates this design in comparison to existing approaches.

Building directly on the theoretical background introduced in the previous section, the goal is twofold: first, to provide intuition for how the proposed experiment was conceived, and second, to present the method with sufficient technical detail to make the implementation clear and reproducible.

\subsubsection{Initialization of Embeddings via a Teacher Model}

The proposed method leverages the model itself as a teacher for the initialization of new embeddings. The key property to be achieved is that the new module integrates as seamlessly as possible into the model from the start, before any fine-tuning on task data is performed. In other words, viewing an LLM as an autoregressive model, for a sequence of tokens the model produces a probability distribution over the next token,
\begin{align}
    P(t_{n+1} \mid t_n, t_{n-1}, t_{n-2}, \ldots, t_0),
\end{align}
but, given an extended vocabulary, concatenated tokens will appear in the sequence, leading to a modified distribution
\begin{align}
    Q(t_{n+1} \mid t_n, t_{(n-1,n-2)}, t_{n-3}, \ldots, t_0),
\end{align}
where $t_{(n-1,n-2)}$ represents a newly introduced token that replaces the pair $(t_{n-1}, t_{n-2})$, as in Table~\ref{tab:awedist_mappings}. Since the underlying text is identical in both cases, the goal is for $P$ and $Q$ to be as similar as possible.

The problem can thus be formulated as finding values for the embedding of $t_{(n-1,n-2)}$ such that the distribution $Q$ induced by the extended tokenizer matches the original distribution $P$ as closely as possible.

\begin{figure}[H]
    \centering
    \hspace{-1cm}
    \makebox[\textwidth]{\includegraphics[width=1.3\textwidth]{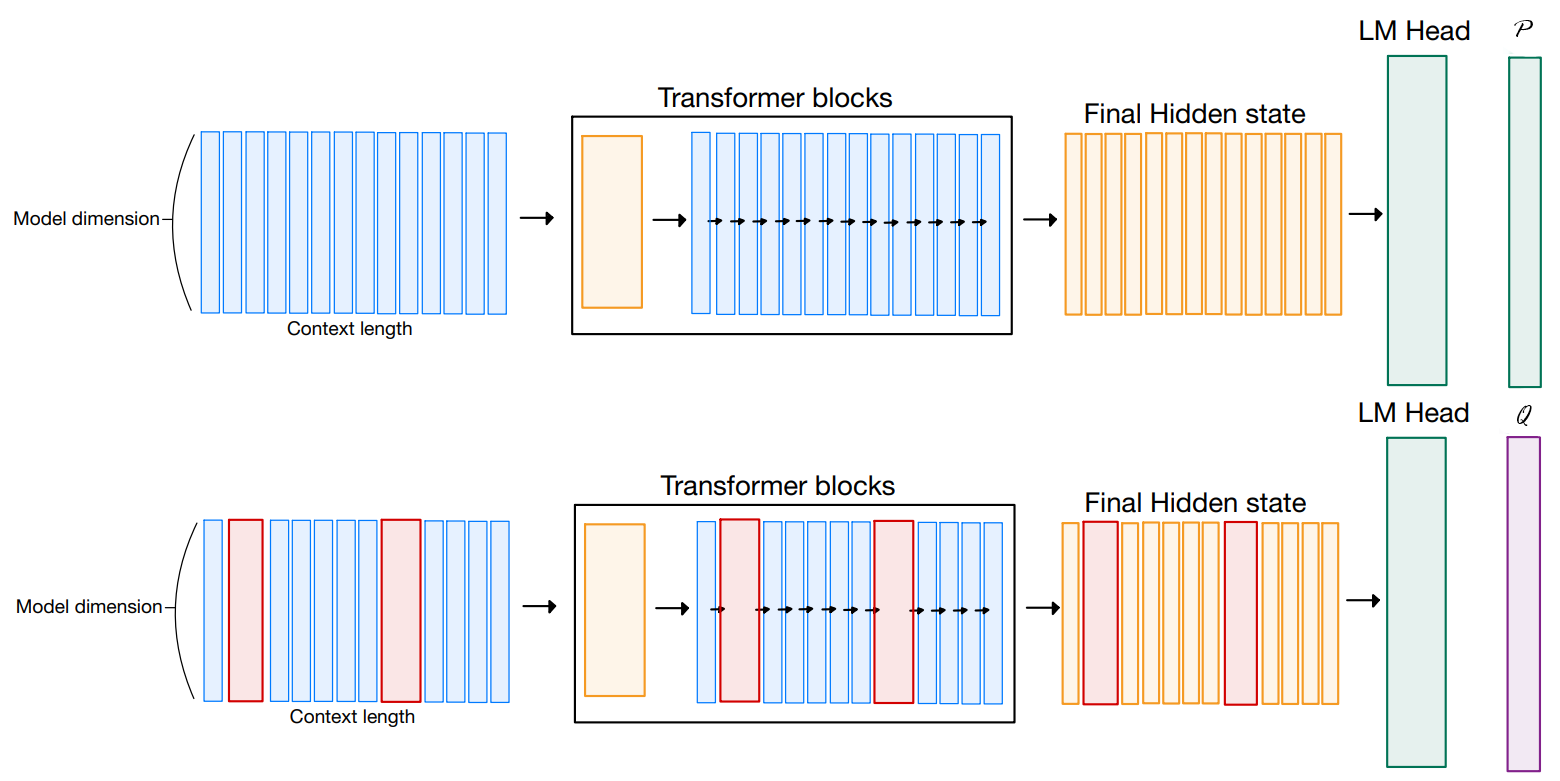}}
    \caption{Illustration of the pipeline for training embeddings via self-distillation. Red bars indicate tokens that are concatenated. The input and internal states are therefore different for the model with and without extended vocabulary. The goal is to find values for the embeddings of these tokens such that KL-divergence is minimized.}
    \label{fig:algo1}
\end{figure}

The process of training embeddings via knowledge distillation, visualized in Figure~\ref{fig:algo1}, uses the model itself to learn new representations for the extended vocabulary that remain aligned with its pretraining; this can be viewed as a form of self-distillation. The procedure is memory-efficient for a teacher–student setup, since only a single model instance needs to be loaded into device memory at any time. The teacher branch uses the original tokenizer, whereas the student branch uses the extended tokenizer. During backpropagation, only the new embeddings are updated, with a training objective that aligns $Q$ to $P$ via KL divergence. When optimizing the embeddings using KL divergence, gradients are taken with respect to the probabilities over the original vocabulary.

Existing methods often propose simple operations such as taking means of embeddings to approximate this alignment. In contrast, the method proposed here initializes embeddings by explicitly training them to minimise the divergence between the two output distributions:

\begin{algorithm}[H]
\caption{Training pipeline for extended-vocabulary input}
\label{alg:embed-init}
\begin{algorithmic}[1]
\State Input text $\xrightarrow{T_{\text{original}}}$ $[t_1, t_2, t_3, \ldots, t_n]$ $\xrightarrow{E_{\text{original}}}$ $[e_1, e_2, e_3, \ldots, e_n]$ $\xrightarrow{\text{LLM}}$ $P$
\State Input text $\xrightarrow{T_{\text{extended}}}$ $[t_1, t_{23}, \ldots, t_n]$ $\xrightarrow{\textcolor{red}{E_{\text{extended}}}}$ $[e_1, \textcolor{red}{e_{23}}, \ldots, e_n]$ $\xrightarrow{\text{LLM}}$ $\textcolor{red}{Q}$ \Comment{$\textcolor{red}{\text{Red}}$ indicates tensors with gradients}
\State $\textcolor{red}{\mathcal{L}} = D_{\text{KL}}(P \,\|\, \textcolor{red}{Q})$ \Comment{KL-divergence loss}
\State Optimize $\textcolor{red}{E_{\text{extended}}}$ parameters: $\theta_{\text{ext}} \leftarrow \theta_{\text{ext}} - \alpha \nabla_{\theta_{\text{ext}}} \textcolor{red}{\mathcal{L}}$ \Comment{Only $\textcolor{red}{E_{\text{extended}}}$ parameters are updated}
\end{algorithmic}
\end{algorithm}

In other words, for an embedding such as $e_{23}$, the KL-divergence loss is used to adjust its values so that the resulting distribution $Q$ is as close as possible to the distribution that would have been produced using the original tokenizer.

The key insight is that the new embeddings are not trained using cross-entropy loss against hard labels, but rather to fit into the existing distributional structure induced by the pretrained model. In an idealised scenario, the KL loss could be driven close to zero without unfreezing any of the original model weights. In practice, this is not achievable, since the self-attention layers are not initially configured to expect concatenations of tokens. Nevertheless, the downstream objective is to make subsequent fine-tuning as easy as possible by starting from embeddings that are already well integrated into the model’s internal representation space.

\subsubsection{Token Alignment}

In theory, the idea is straightforward. In the original setting, given the input text
\[
[\texttt{"import"},\ \texttt{" num"},\ \texttt{"py"},\ \texttt{" "},\ \texttt{"as"},\ \texttt{" "}]
\]
the model produces an output probability distribution for the next token,
\[
P \in \mathbb{R}^{|V|},
\]
where, in this example, the most probable next token is \texttt{"np"}.

Given the new tokenization
\[
[\texttt{"import"},\ \texttt{" numpy"},\ \texttt{" "},\ \texttt{"as"},\ \texttt{" "}],
\]
the goal is to find an embedding for \texttt{"numpy"} such that $D_{\mathrm{KL}}(P \,\|\, Q)$ is minimized, where $Q$ denotes the next-token distribution produced under the extended tokenizer.

\begin{figure}[H]
    \centering
    \hspace{-1cm}
    \makebox[\textwidth]{\includegraphics[width=1.3\textwidth]{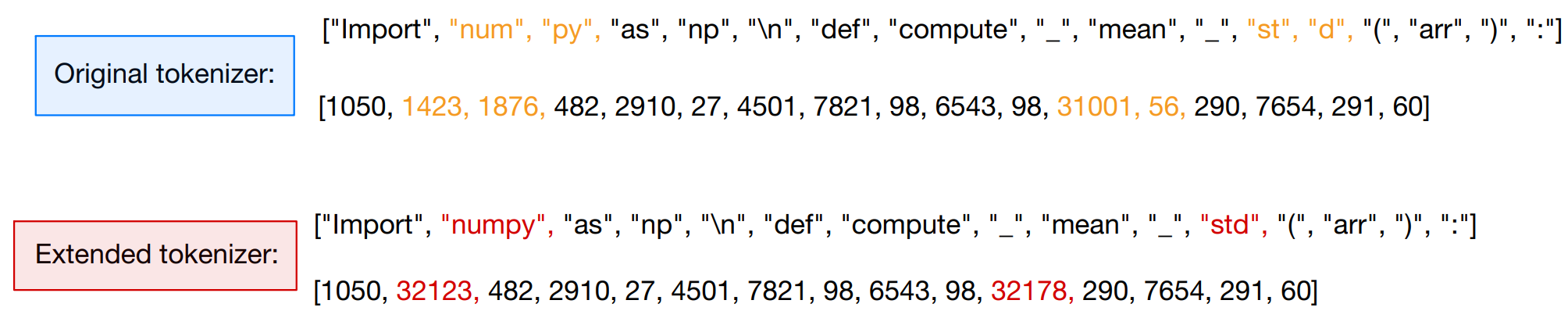}}
    \caption{Adding new tokens creates new tokenization with new indices.}
    \label{fig:tokenization}
\end{figure}

For a single-token example, a direct comparison is trivial, but in practice LLMs are fed full sequences of text and use masked self-attention to train on all tokens at once for efficiency, while still preventing information leakage from future tokens. For longer sequences, there are tokens after \texttt{"np"} as well as many other positions where new tokens appear. Referencing back to Figure~\ref{fig:token1}, only positions corresponding to the same underlying text (i.e., identical tokens or exact word-level alignments) can be meaningfully compared. In that context, the loss is computed only between logits that correspond to the same token in the original vocabulary.

Conceptually, this is the same idea as proposed by AweDist~\cite{dobler2025awedist}, with the key difference that here the loss is computed directly on the output logits (or their induced probabilities) rather than on hidden states. The motivation for choosing KL-divergence on output distributions instead of mean squared error on hidden states is that KL divergence provides a more direct signal for knowledge transfer and is widely adopted in model distillation. The challenge, as pointed out by the authors of AweDist, is that $P$ and $Q$ cannot be compared if they have different dimensionalities. The crucial observation in this work is that the logits (and corresponding probabilities) associated with the extended vocabulary can simply be omitted, retaining a probability distribution over the original vocabulary only. This yields $P, Q \in \mathbb{R}^{|V_{\text{orig}}|}$, making $D_{\mathrm{KL}}(P \,\|\, Q)$ well-defined.

In addition to computing the loss between positions with identical tokenization, a loss term is also computed between the first logits of divergent sequences. In Figure~\ref{fig:tokenization}, this corresponds, for example, to aligning the probabilities for logits associated with \texttt{"num"} and \texttt{"numpy"}, as well as \texttt{"st"} and \texttt{"std"}. This reflects the modelling assumption that, under the new regime, probabilities such as $P(\texttt{"num"}) \approx Q(\texttt{"numpy"})$ should hold whenever these tokens represent the same semantic unit in context.

\begin{figure}[H]
\centering
\scriptsize
\begin{tabular}{c}

\textbf{Original Model Logits} \\
$L_{\text{orig}} \in \mathbb{R}^{17 \times V_{\text{orig}}}$ \\[0.2cm]
\begin{tabular}{|c|c|c|c|c|c|c|c|c|c|c|c|c|c|c|c|c|}
\hline
0 & 1 & 2 & 3 & 4 & 5 & 6 & 7 & 8 & 9 & 10 & 11 & 12 & 13 & 14 & 15 & 16 \\
\hline
\textcolor{blue}{Import} & \textcolor{red}{num} & \textcolor{red}{py} & \textcolor{blue}{as} & \textcolor{blue}{np} & \textcolor{blue}{\textbackslash n} & \textcolor{blue}{def} & \textcolor{blue}{compute} & \textcolor{blue}{\_} & \textcolor{blue}{mean} & \textcolor{blue}{\_} & \textcolor{red}{st} & \textcolor{red}{d} & \textcolor{blue}{(} & \textcolor{blue}{arr} & \textcolor{blue}{)} & \textcolor{blue}{:} \\
\hline
\end{tabular} \\[0.5cm]

\textbf{Extended Model Logits} \\
$L_{\text{ext}} \in \mathbb{R}^{15 \times V_{\text{extended}}}$ \\[0.2cm]
\begin{tabular}{|c|c|c|c|c|c|c|c|c|c|c|c|c|c|c|}
\hline
0 & 1 & 2 & 3 & 4 & 5 & 6 & 7 & 8 & 9 & 10 & 11 & 12 & 13 & 14 \\
\hline
\textcolor{blue}{Import} & \textcolor{red}{numpy} & \textcolor{blue}{as} & \textcolor{blue}{np} & \textcolor{blue}{\textbackslash n} & \textcolor{blue}{def} & \textcolor{blue}{compute} & \textcolor{blue}{\_} & \textcolor{blue}{mean} & \textcolor{blue}{\_} & \textcolor{red}{std} & \textcolor{blue}{(} & \textcolor{blue}{arr} & \textcolor{blue}{)} & \textcolor{blue}{:} \\
\hline
\end{tabular} \\[0.5cm]

$\downarrow$ \textbf{Step 1: Keep similar sequences and first token of divergent sequences} \\[0.2cm]

\begin{tabular}{c}
$L_{\text{orig}}' \in \mathbb{R}^{15 \times V_{\text{orig}}}$ \\[0.1cm]
\begin{tabular}{|c|c|c|c|c|c|c|c|c|c|c|c|c|c|c|}
\hline
0 & 1 & 3 & 4 & 5 & 6 & 7 & 8 & 9 & 10 & 11 & 13 & 14 & 15 & 16 \\
\hline
\textcolor{blue}{Import} & \textcolor{red}{num} & \textcolor{blue}{as} & \textcolor{blue}{np} & \textcolor{blue}{\textbackslash n} & \textcolor{blue}{def} & \textcolor{blue}{compute} & \textcolor{blue}{\_} & \textcolor{blue}{mean} & \textcolor{blue}{\_} & \textcolor{red}{st} & \textcolor{blue}{(} & \textcolor{blue}{arr} & \textcolor{blue}{)} & \textcolor{blue}{:} \\
\hline
\end{tabular} \\[0.3cm]

$L_{\text{ext}}' \in \mathbb{R}^{15 \times V_{\text{extended}}}$ \\[0.1cm]
\begin{tabular}{|c|c|c|c|c|c|c|c|c|c|c|c|c|c|c|}
\hline
0 & 1 & 2 & 3 & 4 & 5 & 6 & 7 & 8 & 9 & 10 & 11 & 12 & 13 & 14 \\
\hline
\textcolor{blue}{Import} & \textcolor{red}{numpy} & \textcolor{blue}{as} & \textcolor{blue}{np} & \textcolor{blue}{\textbackslash n} & \textcolor{blue}{def} & \textcolor{blue}{compute} & \textcolor{blue}{\_} & \textcolor{blue}{mean} & \textcolor{blue}{\_} & \textcolor{red}{std} & \textcolor{blue}{(} & \textcolor{blue}{arr} & \textcolor{blue}{)} & \textcolor{blue}{:} \\
\hline
\end{tabular}
\end{tabular} \\[0.5cm]

$\downarrow$ \textbf{Step 2: Dimension mismatch - Extended model has larger vocabulary} \\[0.3cm]

\begin{tabular}{cc}
\begin{tabular}{c}
$L_{\text{orig}}' \in \mathbb{R}^{15 \times V_{\text{orig}}}$ \\[0.1cm]
\begin{tikzpicture}[scale=1.2]
\draw[thick] (0,0) rectangle (3,2);
\node at (1.5,1) {\tiny $V_{\text{orig}}$};
\node[below] at (1.5,0) {\tiny 15 cols};
\end{tikzpicture}
\end{tabular} &
\begin{tabular}{c}
$L_{\text{ext}}' \in \mathbb{R}^{15 \times V_{\text{extended}}}$ \\[0.1cm]
\begin{tikzpicture}[scale=1.2]
\draw[thick] (0,0) rectangle (3,2);
\draw[thick, red] (0,2) rectangle (3,3);
\node at (1.5,1) {\tiny $V_{\text{orig}}$};
\node[red] at (1.5,2.5) {\tiny $V_{\text{new}}$};
\node[below] at (1.5,0) {\tiny 15 cols};
\end{tikzpicture}
\end{tabular}
\end{tabular} \\[0.5cm]

$\downarrow$ \textbf{Scale back: Remove extended vocabulary rows} \\[0.3cm]

\begin{tabular}{cc}
\begin{tabular}{c}
$P = \text{softmax}(L_{\text{orig}}') \in \mathbb{R}^{15 \times V_{\text{orig}}}$ \\[0.1cm]
\begin{tikzpicture}[scale=1.2]
\draw[thick] (0,0) rectangle (3,2);
\node at (1.5,1) {\tiny $V_{\text{orig}}$};
\node[below] at (1.5,0) {\tiny 15 cols};
\end{tikzpicture}
\end{tabular} &
\begin{tabular}{c}
$Q = \text{softmax}(L_{\text{ext}}'^{\text{scaled}}) \in \mathbb{R}^{15 \times V_{\text{orig}}}$ \\[0.1cm]
\begin{tikzpicture}[scale=1.2]
\draw[thick] (0,0) rectangle (3,2);
\node at (1.5,1) {\tiny $V_{\text{orig}}$};
\node[below] at (1.5,0) {\tiny 15 cols};
\end{tikzpicture}
\end{tabular}
\end{tabular} \\[0.5cm]

$\mathcal{L} = \sum_{i=1}^{15} D_{\text{KL}}(P_i \| Q_i)$ \quad (column-wise KL divergence) \\[0.9cm]

\textbf{Legend:} \\
\textcolor{blue}{Similar tokens} - Same across tokenizations \\
\textcolor{red}{Divergent tokens/vocab} - Different tokenizations or extended vocabulary

\end{tabular}
\caption{Complete loss computation pipeline for embeddings. Step 1 keeps similar sequences and first token of divergent sequences (num from num/py, numpy from numpy, st from st/d, std from std). Step 2 addresses vocabulary size mismatch by removing extended vocabulary rows, resulting in matrices of identical dimensions for column-wise KL divergence computation that allows us to optimize new embeddings but with output over original vocabulary.}
\label{fig:complete_loss_pipeline}
\end{figure}

In short, given an input sequence of length $N$, logits of dimension $|V_{\text{original}}| \times N_{\text{original}}$ are obtained for the non-extended vocabulary model and logits of dimension $|V_{\text{extended}}| \times N_{\text{extended}}$ for the model with the extended vocabulary, where typically $N_{\text{extended}} < N_{\text{original}}$. Referring back to Algorithm~\ref{alg:embed-init}, these logits are two matrices. The first step is to remove the rows corresponding to the new vocabulary from the extended-logit matrix so that the resulting shapes become $|V_{\text{original}}| \times N_{\text{original}}$ and $|V_{\text{original}}| \times N_{\text{extended}}$. 

For training to be valid and consistent with the autoregressive objective, the token IDs must be aligned so that logits corresponding to the same positions in the underlying text are compared. The details of this alignment procedure are given in Algorithm~\ref{alg:token-mapping} in Appendix~C. Using the sequence mappings produced there, columns are removed from the $|V_{\text{original}}| \times N_{\text{extended}}$ matrix so that both matrices are reduced to the shape $|V_{\text{original}}| \times N_{\text{aligned}}$. A column-wise softmax is then applied to each matrix, yielding two probability matrices in which each column represents $P$ and $Q$, respectively, and each column corresponds to the same position in the original string. The KL divergence is computed column-wise on these aligned probability matrices to obtain the final loss value.

\subsubsection{Head}

The head of the LLM predicts the next token given the final hidden state. In the training setting considered here, the final hidden states are denoted by $\mathbf{H} \in \mathbb{R}^{d \times N}$ and the head by a weight matrix $\mathbf{W} \in \mathbb{R}^{|V| \times d}$. The matrix multiplication
\[
    \mathbf{Z} = \mathbf{W}\mathbf{H}
\]
yields logits $\mathbf{Z} \in \mathbb{R}^{|V| \times N}$ that, after applying a column-wise softmax, correspond to next-token probability distributions for each position in the sequence.

To add support for generating tokens from the extended vocabulary, $\mathbf{W}$ is extended by as many additional rows as the vocabulary is expanded with, resulting in logits of shape $|V_{\text{extended}}| \times N$. In practice, this is implemented as a separate module so that it can be trained independently. Since the teacher model does not assign probabilities to the new tokens (they do not exist in its output space), it cannot be used directly to supervise these additional rows in the head. Consequently, the training objective for this module is the standard next-token prediction objective used for LLMs, i.e.\ cross-entropy loss with respect to the ground-truth next token. The exact implementation is as follows:

\begin{algorithm}
\caption{Training pipeline for generating new tokens. Let V be set of vocabulary/tokens (originally 32022 in Deep-seek model), and the F be final hidden states.}
\begin{algorithmic}[1]
\State Input text $\xrightarrow{T_{\text{extended}}}$ $[t_1, t_{23}, ..., t_n] \xrightarrow{E_{\text{extended}}} [e_1, e_{23}, ..., e_n] \xrightarrow{\text{Transformer Blocks}} F$ \Comment{$F \in \mathbb{R}^{d_{\text{model}} \times (n-1)}$}
\State $F \xrightarrow{H_{\text{original}}} L_{\text{original}}$ \Comment{$L_{\text{original}} \in \mathbb{R}^{|V_{\text{original}}| \times (n-1)}$}
\State $F \xrightarrow{\textcolor{red}{H_{\text{extended}}}} \textcolor{red}{L_{\text{new}}}$ \Comment{$L_{\text{extended}} \in \mathbb{R}^{|V_{\text{extended}}| \times (n-1)}$, trainable}
\State Concatenate: $[L_{\text{original}}, \textcolor{red}{L_{\text{new}}}] \xrightarrow{\text{Softmax}} \textcolor{red}{Q}$ \Comment{Final probability distribution}
\State $\textcolor{red}{\mathcal{L}} = \text{CE}(y_{\text{true}}, \textcolor{red}{Q})$ \Comment{Cross-entropy loss}
\State Optimize $\textcolor{red}{H_{\text{extended}}}$: $\theta \leftarrow \theta - \alpha \nabla_{\theta} \textcolor{red}{\mathcal{L}}$ \Comment{Update only trainable parameters}
\end{algorithmic}
\label{alg:head_init}
\end{algorithm}

However, the head module can be initialized in a more informed way than with purely random values. For a new token $t_{\text{new}}$ that is a composition of original tokens $t_1, t_2, \ldots$, the corresponding row for $t_{\text{new}}$ in the extended head is set equal to the row of one of its constituent tokens, denoted here by $t_1$. In natural language terms, if the new token \texttt{"numpy"} is a composition of \texttt{"num"} and \texttt{"py"}, the row of the head corresponding to \texttt{"numpy"} is initialised to be identical to that of \texttt{"num"}. In the pretrained setting, the model is already inclined to predict \texttt{"num"}; by assigning \texttt{"numpy"} the same logit values as \texttt{"num"}, the extended head is initialized in a way that is compatible with the pretrained distribution and minimizes the loss from epoch~0. Concretely, if \texttt{"num"} has ID 1423, the 1423rd row of $\mathbf{W}$ is copied into the row of the extended head that corresponds to the index of \texttt{"numpy"}.

This procedure yields a better-than-random initialization that intuitively improves next-token prediction from the outset and provides a good starting point for subsequent fine-tuning. In the example above, the newly added part of the head only has to learn the nuance between \texttt{" num"} and \texttt{"numpy"}, rather than learning the logits for \texttt{"numpy"} from scratch, which would require more epochs and would typically lead to stronger overfitting on limited data.

It should be emphasized that this heuristic initialization is applied only to the head and is not claimed to be equally effective for embeddings. The reason for this distinction is that the head $\mathbf{X}^{(N)}$ acts as a single linear transformation on the final hidden states $\mathbf{H}$; setting specific rows in the extended head directly enforces a particular behavior at the level of output logits, which is relatively tractable. In contrast, setting input embeddings to a linear combination of constituent embeddings provides no such guarantee: the subsequent processing by multiple self-attention and feed-forward blocks is highly non-linear, and all embeddings and hidden states interact in complex ways that are effectively intractable to reason about analytically. This motivates the choice to \emph{optimize} the new embeddings via KL-based self-distillation, while \emph{initializing} the head heuristically and training it with the standard cross-entropy objective.

\subsection{Experimental Setup}

This subsection describes the overall experimental configuration used to evaluate the proposed method. It outlines the details of how the experiments were run, with the goal that the results can be reproduced as closely as possible.

\subsubsection{Training of Head and Embedding}

The training process for the head and embeddings consisted of four methods run in parallel to compare the proposed approach with previous baselines:

\begin{enumerate}
    \item \textbf{Random + Cross-Entropy Baseline}: Embeddings and head were initialized randomly and trained using the standard next-token prediction objective with cross-entropy loss.
    
    \item \textbf{Mean + Cross-Entropy}: Embeddings and head were initialized using mean initialization of embeddings (averaging constituent subtoken embeddings) and trained from this initialization with cross-entropy loss.
    
    \item \textbf{Random + Proposed Method}: Embeddings were initialized randomly and optimized using the proposed knowledge-distillation method with KL-divergence loss, while the head was trained with cross-entropy loss.
    
    \item \textbf{Mean + Proposed Method}: Embeddings were initialized using mean initialization and then optimized using the proposed knowledge-distillation method with KL-divergence loss, while the head was trained with cross-entropy loss.
\end{enumerate}

It is worth noting that, although Algorithms~\ref{alg:embed-init} and~\ref{alg:head_init} are described as two separate phases, both $\textcolor{red}{E_{\text{extended}}}$ and $\textcolor{red}{H_{\text{extended}}}$ were trained simultaneously for all four methods, but with different optimizers assigned to the respective modules. While all four methods used cross-entropy loss for $\textcolor{red}{H_{\text{extended}}}$, Methods~3 and~4 additionally used KL-divergence loss for $\textcolor{red}{E_{\text{extended}}}$. This setup is feasible because the two optimizers are attached to disjoint modules and operate on separate computational graphs. In practical terms, the gradients obtained via the chain rule for the parameters of the head are unaffected by the embeddings, and vice versa. In other words, optimizing the head for the extended vocabulary has no impact on the gradients of the embeddings (which are optimized with respect to the original vocabulary), and optimizing the embeddings does not affect the gradients of the extended head. This dual-optimizer setup with separate backward passes effectively halves the wall-clock time required for this phase of the training pipeline.

All four configurations were trained for a total of 12~epochs, until the decrease in loss plateaued, indicating convergence of the new token representations. The learning rate was set to $4.2 \times 10^{-4}$ with a linear warmup followed by linear decay using a 10/90 split between warmup and decay. Training was first run for 8~epochs and then continued for an additional 4~epochs, which effectively corresponds to a warm restart of the learning-rate schedule.

\subsubsection{Training of Whole Head/Embedding}

In the next step, a fifth model was introduced for subsequent training, namely a model with randomly initialized $\textcolor{red}{E_{\text{extended}}}$ and $\textcolor{red}{H_{\text{extended}}}$ that had not undergone any optimization as described in the previous subsection. This additional configuration was included to provide a further comparison point in the experimental results and to assess the importance of the initial optimization phase versus direct full-vocabulary training.

All five configurations were then trained by unfreezing the entire embedding and head layers and running an additional 4~epochs of training. It is necessary not only to train the newly added head rows and embeddings (as in the previous phase), but also to perform several epochs of training over the entire vocabulary. This is because tokens such as \texttt{"num"} and \texttt{"py"} will appear in altered contexts once a new token such as \texttt{"numpy"} has been introduced. The existing embeddings therefore need to adapt to the presence of merged tokens, which modifies the distributional patterns on which they were originally trained.

Referring back to Figure~\ref{fig:complete_pipeline}, this part of the process corresponds to the actual vocabulary expansion of the LLM, where only the input and output embedding layers of the model are updated. As discussed in Section~\ref{sec:disc_emb}, modifying only these parts of the model is not sufficient to obtain a fully functional system: the self-attention layers are pretrained to process particular token patterns and keywords that are now represented by different tokens. For this reason, an additional fine-tuning phase on the full model is required.

\subsubsection{Full Finetuning with LoRA}

As a final step, all five model versions were fine-tuned using LoRA on all transformer blocks and all modules within those blocks (all attention matrices and linear layers). LoRA was chosen due to its regularizing effect during fine-tuning, which helps mitigate catastrophic forgetting of pretrained knowledge while adapting to new domain-specific patterns~\cite{biderman2024lora,ren2024analyzing}. The low-rank constraint inherent in LoRA’s decomposition acts as an implicit regularizer, reducing overfitting to the target domain while preserving the model’s general capabilities learned during pretraining. By training only a small number of additional parameters, LoRA constrains the fine-tuned model from diverging significantly from the base model, which is particularly beneficial in continual-learning scenarios where specialization in new domains can otherwise come at the expense of base-model capabilities~\cite{hu2021lora}.

\subsubsection{Full Sequential Finetuning}

In addition to LoRA-based fine-tuning, a full-parameter fine-tuning phase was carried out in which the transformer blocks were completely unfrozen. This phase was conducted for three model variants deemed most suitable for further evaluation:
\begin{itemize}
    \item the base model without extended vocabulary, serving as a baseline. Since fine-tuning a pretrained model typically leads to a slight degradation in general performance, fine-tuning the base model itself provides a reference that controls for forgetting effects;
    \item the model with embeddings optimized using the conventional cross-entropy objective (i.e., the standard way of fine-tuning an LLM with vocabulary expansion); 
    \item the model with embeddings initialized and optimized using the proposed teacher-based (self-distillation) method.
\end{itemize}

As a starting point for full fine-tuning, the model weights were taken from the state obtained after the embedding and head layers had been unfrozen and trained. At that point, Phase~1 (initialization and optimization of embeddings and head) had been completed, followed by 4~full epochs of training with the entire embedding and head layers unfrozen.

The selected models were then trained sequentially with partial unfreezing: first, the initial 16 transformer blocks were unfrozen and trained for 1~epoch, after which the final 16 transformer blocks were unfrozen and trained for an additional epoch. The learning rate was set to $4.2 \times 10^{-5}$, matching the rate reported for full fine-tuning of DeepSeek Coder~6.7B in its final training phase. The same learning rate was chosen in line with the recommendation that fine-tuning should use a learning rate close to that employed during pretraining~\cite{parmar2024reuse}.

\section{Evaluation}

For evaluation, loss was not used as the primary metric, since the setting is highly overparameterized and training loss does not necessarily correlate well with downstream performance. Instead, standardized benchmarks for code generation were employed. In particular, BigCodeBench~\cite{zhuo2024bigcodebench} and DS-1000~\cite{pmlr-v202-lai23b} were selected, as they are specifically designed for code models and serve as the main evaluation metrics in this work.

BigCodeBench is a benchmark consisting of function-level programming tasks that require invoking multiple function calls from popular libraries. Unlike simple algorithmic benchmarks, it focuses on realistic programming scenarios with complex instructions. DS-1000 is a code-generation benchmark with 1000 data-science problems spanning seven Python libraries (NumPy, Pandas, etc.). It focuses on instruction/code-completion problems where models generate Python code that is executed in the same environment used for inference and evaluated against assertions and test cases for functional correctness. These benchmarks were chosen because they offer a realistically high density of tokens from the extended vocabulary.

Both benchmarks span diverse programming domains and follow similar evaluation procedures: models generate code based on prompts, the generated code is executed, and the results are checked against test cases to verify correctness. For reference, GPT-4o achieves 51.1\,\% on BigCodeBench-Instruct~\cite{hf_bigcodebench_blog_2024}.

For the purposes of this thesis, both datasets were cleaned by running the provided reference solutions and removing instances that produced incorrect answers due to benchmark errors or environment issues. This filtering resulted in 1022 problems from BigCodeBench and 929 problems from DS-1000 being retained for evaluation.

\section{Results}

This section presents the results of the experiments. It is divided into two main parts: the training behaviour, which reports the observed loss curves during optimization, and the evaluation results on external benchmarks.

\subsection{Training}

This subsection reports the observed training losses. Although loss values alone do not fully characterize model performance, in this setting they provide important insight into the training dynamics under the proposed vocabulary-expansion regime and are valuable in their own right. The loss curves also support more grounded conclusions in the subsequent analysis.

\subsubsection{Phase 1: New Module Training}

The first part of the results concerns the loss curves from training only the newly introduced modules. Two types of initialization (mean and random) and two training objectives for the embeddings (standard cross-entropy vs.\ KL-based teacher training) were combined, yielding four methods in total. The head was trained with cross-entropy loss in all four configurations. For the proposed method, two optimizers were used in parallel: one with a KL-divergence loss for the embeddings and one with a cross-entropy loss for the head. Since the corresponding computational graphs do not intersect, both losses could be optimized simultaneously without mutual interference, improving training efficiency. In the reference cross-entropy setting, a single optimizer was applied to both embeddings and head, but with two separate backward passes to maintain numerical stability.

\begin{figure}[H]
\hspace*{-2cm}
\begin{tikzpicture}
\begin{axis}[
    xlabel=Epoch,
    ylabel=Cross-Entropy Loss,
    width=1.2\textwidth,
    height=9cm,
    legend pos=north east,
    grid=major,
    ymin=0.5,
    ymax=0.9,
    xmin=0,
    xmax=12,
    cycle list={
        {red, solid},
        {green, solid},
        {blue, solid},
        {orange, solid},
        {black, solid}
    }
]
% Training losses (solid lines with markers)
\addplot [red, solid, mark=*, mark options={solid, fill=red}] table [x index=0, y index=1, col sep=comma] {data/phase1/train_ce_losses.csv};
\addlegendentry{Mean init, standard next-token prediction}
\addplot [green, solid, mark=*, mark options={solid, fill=green}] table [x index=0, y index=3, col sep=comma] {data/phase1/train_ce_losses.csv};
\addlegendentry{Random init, standard next-token prediction}
\addplot [blue, solid, mark=*, mark options={solid, fill=blue}] table [x index=0, y index=2, col sep=comma] {data/phase1/train_ce_losses.csv};
\addlegendentry{Mean init, with teacher model}
\addplot [orange, solid, mark=*, mark options={solid, fill=orange}] table [x index=0, y index=4, col sep=comma] {data/phase1/train_ce_losses.csv};
\addlegendentry{Random init, with teacher model}
% Validation losses (dashed lines with markers)
\addplot [red, dashed, mark=*, mark options={solid, fill=red}] table [x index=0, y index=1, col sep=comma] {data/phase1/val_ce_losses.csv};
\addplot [green, dashed, mark=*, mark options={solid, fill=green}] table [x index=0, y index=3, col sep=comma] {data/phase1/val_ce_losses.csv};
\addplot [blue, dashed, mark=*, mark options={solid, fill=blue}] table [x index=0, y index=2, col sep=comma] {data/phase1/val_ce_losses.csv};
\addplot [orange, dashed, mark=*, mark options={solid, fill=orange}] table [x index=0, y index=4, col sep=comma] {data/phase1/val_ce_losses.csv};
% Dummy plot for validation legend entry
\addplot [black, dashed, opacity=0] coordinates {(0,0)};
\addlegendentry{Validation (dashed line)}
\end{axis}
\end{tikzpicture}
\caption{Cross-entropy loss during Phase~1 training.}
\label{fig:phase1_ce_loss}
\end{figure}

The final cross-entropy loss was observed to be largely independent of the embedding-training objective, but clearly affected by the choice of head initialization, as shown in Figure~\ref{fig:phase1_ce_loss}. This behaviour is consistent with a highly non-convex optimization landscape. It can also be observed that the validation loss is slightly lower than the training loss, indicating that the validation subset is somewhat more favourable to the model than the training subset (despite being drawn from the same dataset) and that the setting remains sufficiently underparameterized to avoid overfitting in this phase.

\newpage

\begin{figure}[H]
\hspace*{-2cm}
\begin{tikzpicture}
\begin{axis}[
    xlabel=Epoch,
    ylabel=KL Divergence Loss,
    width=1.2\textwidth,
    height=9cm,
    legend pos=north east,
    grid=major,
    ymin=0.07,
    xmin=0,
    xmax=12,
    cycle list={
        {red, solid},
        {green, solid},
        {blue, solid},
        {orange, solid},
        {black, solid}
    }
]
% Training losses (solid lines with markers)
\addplot [red, solid, mark=*, mark options={solid, fill=red}] table [x index=0, y index=1, col sep=comma] {data/phase1/train_kl_losses.csv};
\addlegendentry{Mean init, Standard next token prediction}
\addplot [green, solid, mark=*, mark options={solid, fill=green}] table [x index=0, y index=3, col sep=comma] {data/phase1/train_kl_losses.csv};
\addlegendentry{Random init, Standard next token prediction}
\addplot [blue, solid, mark=*, mark options={solid, fill=blue}] table [x index=0, y index=2, col sep=comma] {data/phase1/train_kl_losses.csv};
\addlegendentry{Mean init, With teacher model}
\addplot [orange, solid, mark=*, mark options={solid, fill=orange}] table [x index=0, y index=4, col sep=comma] {data/phase1/train_kl_losses.csv};
\addlegendentry{Random init, With teacher model}
% Validation losses (dashed lines with markers)
\addplot [red, dashed, mark=*, mark options={solid, fill=red}] table [x index=0, y index=1, col sep=comma] {data/phase1/val_kl_losses.csv};
\addplot [green, dashed, mark=*, mark options={solid, fill=green}] table [x index=0, y index=3, col sep=comma] {data/phase1/val_kl_losses.csv};
\addplot [blue, dashed, mark=*, mark options={solid, fill=blue}] table [x index=0, y index=2, col sep=comma] {data/phase1/val_kl_losses.csv};
\addplot [orange, dashed, mark=*, mark options={solid, fill=orange}] table [x index=0, y index=4, col sep=comma] {data/phase1/val_kl_losses.csv};
% Dummy plot for validation legend entry
\addplot [black, dashed, opacity=0] coordinates {(0,0)};
\addlegendentry{Validation (dashed line)}
\end{axis}
\end{tikzpicture}
\caption{KL divergence loss during Phase 1 training.}
\label{fig:phase1_kl_loss}
\end{figure}

Interestingly, the KL-divergence loss obtained when training embeddings solely with the cross-entropy objective (red and green curves in Figure~\ref{fig:phase1_kl_loss}) appears largely uncorrelated with the KL-divergence loss optimized under the teacher objective (blue and orange curves). For the mean initialization in particular (red), the KL loss even increases over epochs. This behaviour suggests a misalignment between the standard next-token cross-entropy objective and the KL-divergence objective used for self-distillation, with potential downstream consequences.

It can also be observed that the KL-divergence \emph{training} loss for the \textcolor{orange}{random initialization with teacher training} is higher than for the \textcolor{blue}{mean initialization with teacher training}, whereas the opposite holds for the corresponding \emph{validation} losses. This pattern indicates that the heuristic (mean-based) initialization may have a regularizing effect on the optimization landscape, leading to comparatively better generalization despite achieving a lower training KL loss.

Taken together with the cross-entropy results in Figure~\ref{fig:phase1_ce_loss}, the picture is consistent: cross-entropy loss is primarily affected by the choice of initialization, with lower loss observed for the heuristic initialization already before optimization, again pointing to a highly non-convex optimization landscape. In contrast, the KL-divergence loss decreases only when it is explicitly used as a training objective (blue and orange), and increases when it is not (red and green). This behaviour suggests that, in this setting, pure next-token prediction pushes the model away from the original pretrained distribution measured by KL divergence. Up to this phase, the original model weights remain frozen; only the newly added parameters have been updated. The next steps therefore involve gradually unfreezing layers of the original model to allow the remainder of the network to adapt to the extended vocabulary.

\newpage

\subsubsection{Phase 2: Full Finetuning with LoRA}

\begin{figure}[htbp]
\hspace*{-2cm}
\begin{tikzpicture}
\begin{axis}[
   xlabel=Epoch,
   ylabel=Loss,
   width=1.2\textwidth,
   height=9cm,
   legend pos=north east,
   grid=major,
   xmin=0.0,
   xmax=8.6,
   ymax=0.9,
   ymin=0.4,
   xtick={0,1,2,3,4,5,6,7,8},
   cycle list={
       {red, solid},
       {green, solid},
       {blue, solid},
       {orange, solid},
       {purple, solid},
       {black, solid}
   }
]
% Training losses (solid lines with markers)
\addplot [red, solid, mark=*, mark options={solid, fill=red}] table [x index=0, y index=1, col sep=comma] {data/phase2/finetune_train_losses.csv};
\addlegendentry{Mean init, Standard next token prediction}
\addplot [green, solid, mark=*, mark options={solid, fill=green}] table [x index=0, y index=2, col sep=comma] {data/phase2/finetune_train_losses.csv};
\addlegendentry{Random init, Standard next token prediction}
\addplot [blue, solid, mark=*, mark options={solid, fill=blue}] table [x index=0, y index=3, col sep=comma] {data/phase2/finetune_train_losses.csv};
\addlegendentry{Mean init, With teacher model}
\addplot [orange, solid, mark=*, mark options={solid, fill=orange}] table [x index=0, y index=4, col sep=comma] {data/phase2/finetune_train_losses.csv};
\addlegendentry{Random init, With teacher model}
\addplot [purple, solid, mark=*, mark options={solid, fill=purple}] table [x index=0, y index=5, col sep=comma] {data/phase2/finetune_train_losses.csv};
\addlegendentry{Non-optimized embeddings+head baseline}
% Validation losses (dashed lines with markers)
\addplot [red, dashed, mark=*, mark options={solid, fill=red}] table [x index=0, y index=1, col sep=comma] {data/phase2/finetune_val_losses.csv};
\addplot [green, dashed, mark=*, mark options={solid, fill=green}] table [x index=0, y index=2, col sep=comma] {data/phase2/finetune_val_losses.csv};
\addplot [blue, dashed, mark=*, mark options={solid, fill=blue}] table [x index=0, y index=3, col sep=comma] {data/phase2/finetune_val_losses.csv};
\addplot [orange, dashed, mark=*, mark options={solid, fill=orange}] table [x index=0, y index=4, col sep=comma] {data/phase2/finetune_val_losses.csv};
\addplot [purple, dashed, mark=*, mark options={solid, fill=purple}] table [x index=0, y index=5, col sep=comma] {data/phase2/finetune_val_losses.csv};
% Dummy plot for validation legend entry
\addplot [black, dashed, opacity=0] coordinates {(0,0)};
\addlegendentry{Validation (dashed line)}
% Vertical dotted line at x=4
\addplot [black, dotted, line width=1pt] coordinates {(4,0.4) (4,0.9)};
\end{axis}
\end{tikzpicture}
\caption{Losses for finetuning of full LLM. First 4 epochs are head+embeddings only and epoch 5 to 8 is LoRA with rank 128 on all 32 transformer blocks. Learning rate of 2.2e-4 with linear warmup over first epoch.}
\label{fig:finetune_loss}
\end{figure}

The loss was observed to converge, with the final value depending on whether random or heuristic initialization was used, consistent with the earlier phases. It is also worth noting that, while the training loss naturally became lower than the validation loss, the validation loss decreased almost monotonically across epochs. With a constant learning rate, the additional decrease after epoch~8 was negligible, indicating convergence.

This behaviour suggests that there may be headroom for increasing the LoRA rank to further improve performance. At the same time, maintaining a relatively regularized model is beneficial in the present setting, as it allows the use of identical training schedules for all model variants without the need for early stopping. This, in turn, helps isolate the effects of different initialization schemes from other confounding factors in the experimental setup.

\newpage

\subsubsection{Phase 2: Full Sequential Finetuning}

\begin{figure}[htbp]
\hspace*{-2cm}
\begin{tikzpicture}
\begin{axis}[
    xlabel=Epoch,
    ylabel=Loss,
    width=1.2\textwidth,
    height=9cm,
    legend pos=north east,
    grid=major,
    xmin=0,
    xmax=6.5,
    ymin=0.35,
    xtick={0,1,2,3,4,5,6},
    cycle list={
        {red, solid},
        {blue, solid},
        {green, solid},
        {black, solid}
    }
]
% Training losses (solid lines with markers)
\addplot [red, solid, mark=*, mark options={solid, fill=red}] coordinates {
    (1,0.5988806884765625) (2,0.57897607421875) (3,0.5692543212890625) (4,0.5625922607421875) (5,0.5375560180664063) (6,0.3880338134765625)
};
\addlegendentry{Cross entropy optimized (train)}
\addplot [blue, solid, mark=*, mark options={solid, fill=blue}] coordinates {
    (1,0.6237598388671876) (2,0.5933748291015625) (3,0.580024462890625) (4,0.571891796875) (5,0.5372623779296875) (6,0.38918896789550783)
};
\addlegendentry{Teacher trained model (train)}
\addplot [green, solid, mark=*, mark options={solid, fill=green}] coordinates {
    (1,0.5071235168457031) (2,0.3641757652282715)
};
\addlegendentry{Base model (train)}
% Validation losses (dashed lines with markers)
\addplot [red, dashed, mark=*, mark options={solid, fill=red}] coordinates {
    (0,0.6280281982421875) (1,0.594512451171875) (2,0.583063720703125) (3,0.57691259765625) (4,0.572529296875) (5,0.5312041625976562) (6,0.5494957733154296)
};
\addplot [blue, dashed, mark=*, mark options={solid, fill=blue}] coordinates {
    (0,0.6680211181640625) (1,0.611709716796875) (2,0.5949635009765625) (3,0.58625) (4,0.5810255126953126) (5,0.5317838745117187) (6,0.5475254211425781)
};
\addplot [green, dashed, mark=*, mark options={solid, fill=green}] coordinates {
    (0,0.5536316528320312) (1,0.4948560333251953) (2,0.5149693984985352)
};
% Dummy plot for validation legend entry (black instead of red)
\addplot [black, dashed] coordinates {(0,0.35) (0.1,0.35)};
\addlegendentry{Validation (dashed line)}
% Vertical dotted line at x=4
\addplot [black, dotted, line width=1pt] coordinates {(4,0.35) (4,0.7)};
\end{axis}
\end{tikzpicture}
\caption{Full sequential finetuning results. Note that \textcolor{green}{base model} when unfreezing all weights only had undergone one epoch before validation loss started to increase.}
\label{fig:sequential_finetune}
\end{figure}

The first four epochs for the models with extended vocabulary correspond to those shown in Figure~\ref{fig:finetune_loss}, where full fine-tuning of the embedding layers is performed. In epoch~5, the first 16 transformer blocks are unfrozen, and in epoch~6, the remaining 16 transformer blocks are unfrozen. Once the final 16 blocks are unfrozen, the training loss decreases sharply while the validation loss increases, indicating the onset of overfitting. Training was therefore stopped at this point to prevent further divergence between training and validation behaviour. 

This can be compared with Figure~\ref{fig:finetune_loss}, where the regularizing effect of LoRA prevents such a divergence between training and validation losses. In that figure, the green line represents full fine-tuning of the base model on the training data without any vocabulary expansion.

\subsection{Evaluation}

The results from the two benchmarks differ in several respects but are consistent in one key aspect: the proposed method, when combined with heuristic initialization, outperforms all other evaluated models, including the original DeepSeek checkpoint without vocabulary expansion. BigCodeBench focuses on generating entire functions or scripts, whereas DS-1000 primarily targets code-completion style tasks, where the model is asked to finish partially written code and typically produces fewer output tokens. These benchmarks therefore probe different aspects of model performance: extended multi-step generation versus next-token prediction in a highly constrained local context. The implications of these differences for the observed results are discussed in the following section.

\begin{table}[htbp]
\centering
\begin{adjustbox}{width=1.2\textwidth,center}
\begin{tabular}{l|cc|cc|cc}
\hline
\textbf{Model} & \multicolumn{2}{c|}{\textbf{BigCodeBench}} & \multicolumn{2}{c|}{\textbf{DS-1000}} & \multicolumn{2}{c}{\textbf{Combined}} \\
& \textbf{Pass (\%)} & \textbf{Passed/Total} & \textbf{Pass (\%)} & \textbf{Passed/Total} & \textbf{Passed/Total} & \textbf{P-value} \\
\hline
\textbf{Base Model} & & & & & & \\
Base model (no vocabulary expansion) & 20.84 & 213/1022 & 31.75 & 295/929 & 508/1951 & 0.084 \\
\hline
\textbf{LoRA Finetuning} & & & & & & \\
Random init, non-optimized initialization & 19.86 & 203/1022 & 31.11 & 289/929 & 492/1951 & 0.021 \\
Random init, optimized with CE & 16.93 & 173/1022 & 33.80 & 314/929 & 487/1951 & 0.013 \\
Mean init, optimized with CE & 13.89 & 142/1022 & 34.34 & 319/929 & 461/1951 & 0.001 \\
Random init, optimized with KL+CE & 12.62 & 129/1022 & 33.80 & 314/929 & 443/1951 & < 0.001 \\
\textbf{Mean init, optimized with KL+CE (ours)} & \textbf{21.14} & \textbf{216/1022} & \textbf{36.60} & \textbf{340/929} & \textbf{556/1951} & \textbf{reference} \\
\hline
\textbf{Full Finetuning} & & & & & & \\
Base model finetuned (no vocabulary expansion) & 20.16 & 206/1022 & 34.55 & 321/929 & 527/1951 & 0.300 \\
Mean init, optimized with CE & 17.81 & 182/1022 & 32.72 & 304/929 & 486/1951 & 0.011 \\
Mean init, optimized with KL+CE (ours) & 18.79 & 192/1022 & 31.65 & 294/929 & 486/1951 & 0.011 \\
\hline
\end{tabular}
\end{adjustbox}
\caption{Evaluation results across BigCodeBench and DS-1000 benchmarks with statistical significance testing.}
\label{tab:combined_results}
\end{table}

From Table~\ref{tab:combined_results} it can be observed that, for LoRA fine-tuning, the more random initialization with cross-entropy–trained embeddings performs comparatively better on BigCodeBench, whereas the same type of initialization performs substantially worse under the KL+CE training regime. For DS-1000, the relative ranking of methods differs more markedly, reflecting both the nature of the benchmark (code completion rather than long-form generation) and the differing distributions of problem types.

All p-values are computed against the best-performing configuration (Mean init, optimized with KL+CE under LoRA fine-tuning) using two-proportion z-tests on the combined success counts. For all other models that underwent vocabulary expansion, this best-performing variant outperforms the alternatives by a statistically significant margin, as indicated by the reported p-values.

\newpage

\section{Discussion}

The following section analyzes the results and explains why the observed behaviours arise, while also providing insight into the training process. It further connects these findings back to the original research questions and methodological choices of the thesis.

\subsection{Phase 1: Training Only New Embeddings}\label{sec:disc_emb}

Much of the literature on vocabulary expansion, such as AweDist~\cite{dobler2025awedist}, focuses on optimizing only the input and/or output embeddings. However, this approach ignores the fact that the transformer blocks have learned to process many words as compositions of multiple tokens (and corresponding hidden states). When a word such as \texttt{"fibonacci"} changes from a sequence of subtokens (\texttt{"fib"}, \texttt{"ona"}, \texttt{"cci"}) into a single token, the self-attention layers must reinterpret a representation that previously consisted of several positions in the sequence. This shift can produce artefacts: the model may learn a reasonable semantic representation for the new embedding, yet still generate syntactically or lexically odd outputs because the internal transformer layers do not yet know how to handle the new token in context.

\begin{lstlisting}[language=Python, basicstyle=\ttfamily, escapeinside={(*@}{@*)}]
Instruction: Write a Python function that 
calculates Fibonacci sequence:
Response: def fibonacci(*@\textcolor{red}{onacci}@*)(n):
    if n == 0:
        return 0
    elif n == 1:
        return 1
    else:
        return fibonacci(n-1) + fibonacci(n-2)
\end{lstlisting}

In this example, the model produces \texttt{"onacci"} after \texttt{"fibonacci"} because, in the original tokenizer, the first token of the word is \texttt{"fib"}. This behaviour becomes especially problematic in code, where exact syntax matters. At the same time, it shows that the training process does teach the model to interpret the new embeddings to some extent; the remaining problem lies in how the internal layers process them. For this particular example, the embeddings were initialized using the proposed KL-based method.

When the embeddings are instead trained with the standard cross-entropy objective, the model produces the following answer:

\begin{lstlisting}[language=Python, basicstyle=\ttfamily, escapeinside={(*@}{@*)}]
Response: def fibonacci(n):
    if n == 0:
        return 0
    elif n == 1:
        return 1
    else:
        return fibonacci(n-1) + fibonacci(n-2)
\end{lstlisting}

This version is syntactically correct and yields a lower next-token cross-entropy loss for this specific task, so from the perspective of pure next-token prediction, the conventional training objective performs better. However, after full fine-tuning, the benchmark results favour the KL-based method.

Appendix \ref{sec:mechanistic} offers a mechanistic hint as to why this happens. Under cross-entropy training, the new embedding for \texttt{"fibonacci"} aligns more closely in vector space with the last composite subtoken (e.g.\ \texttt{"cci"}), whereas KL-divergence training preserves more information from the earlier subtokens as well. As a result, the model tends to interpret the single token \texttt{"fibonacci"} more like the full word, rather than collapsing its representation towards a single suffix-like component. This distinction matters more in complex contexts than in the simple function above.

Overall, this behaviour highlights a tension between optimizing for immediate next-token prediction and improving the model’s global understanding after vocabulary expansion. It also underlines the limitations of restricting adaptation to embeddings alone: without further fine-tuning of the transformer blocks, the internal computation remains partially misaligned with the new tokenization scheme.

\subsection{Phase 2: Finetuning and Evaluation}

The key finding in this phase is that the proposed method outperforms all other configurations when fine-tuned with LoRA, with a margin that is statistically significant. The most promising—and somewhat surprising—result is that the model even surpasses the base model when optimized with LoRA and initialized heuristically. At the same time, the large variance in outcomes across different settings highlights how strongly initialization influences downstream performance.

\subsubsection{Experimental Setup and Formatting}

Model performance depends heavily on the exact experimental setup. Even though the benchmarks are standardized, there is considerable flexibility in how prompts are tokenized (e.g.\ where special tokens are inserted) and how questions are formatted using a chat template, which itself can vary between implementations. In this experiment, all models use the same chat template, matching the format employed for inference and pretraining, as well as for the fine-tuning carried out here. Concretely, prompts are formatted as:

\begin{verbatim}
<BOS> ### Instruction:
Query

### Answer:
```python
import package

def task_func(argument)
\end{verbatim}

In this setup, the beginning of the code appears in the answer block rather than being included in the instruction block. This choice turns out to be important for interpreting the model’s behaviour in the subsequent analysis.

\subsubsection{BigCodeBench}

BigCodeBench involves the generation of whole functions based on instructions and a few lines of starting code, with results evaluated via one-shot tests. Notably, superior results are observed for heuristic initialization with KL-divergence trained embeddings, while better performance is obtained from randomly initialized embeddings when trained using cross entropy. While the exact causes are complex, hypotheses can be formed based on observed responses and mechanistic interpretation.

The most common failure mode occurs when the model does not continue generating code but instead produces natural language. This is presumed to be caused by the dataset, which frequently includes explanations before markdown code, or sometimes only explanations. Consequently, a bias against direct code generation is formed, leading to a failure to follow the given format when prompted.

It is observed that only randomly initialized KL-divergence trained embeddings fail at relevant next-token prediction. When a sequence contains a small number of new tokens, embeddings are trained not only for immediate KL-divergence loss but, given that gradients flow back across the whole sequence, to provide relevant context for all future tokens. This leads to the behavior observed in Figure~\ref{fig:embedding_similarity}, where values associated with the first composite embedding are adopted to maximize future context. While beneficial for minimizing KL-loss across a sequence, this mechanism can fail during code completion tasks like BigCodeBench, resulting in natural language generation instead of next-token prediction. Conversely, these errors are mitigated when using heuristic initialization, as ``semantic information'' is inherited from sub-embeddings (e.g., forming $e_{fibonacci}$ from the mean of $e_{fib}$, $e_{ona}$, $e_{cci}$) and only adjusted to conform to the pre-trained setting. It was found that most embedding values remained stable during KL-divergence training, though the loss decreases subtly (Figure~\ref{fig:phase1_kl_loss}). This suggests KL-divergence alignment involves changing few indices to fit the pre-trained setting while preserving conceptual knowledge.

This raises the question of why the opposite result is shown for CE-optimized embeddings. Here, failures occurred $\approx$ 50\% of the time due to the generation of natural language instead of code continuation. In an autoregressive sequence, competing signals exist regarding the next action. In this case, the signal to provide an explanation appears stronger than the signal for code completion, causing the function to be interpreted as complete even if code continuation is syntactically more correct.

Example:
\begin{verbatim}
<BOS> ### Instruction:
Generates a CSV file containing simulated data for 100 people, 
including name, age, height, and weight. It also calculates and 
appends the average age, height, and weight at the end of the file. 
The function should output with: str: The path of the created CSV file. 
You should write self-contained code starting with:

### Answer:
```python
import os
import csv
import random
from statistics import mean

# Constants
COLUMNS = ['Name', 'Age', 'Height', 'Weight']
PEOPLE_COUNT = 100

def task_func(filename):
\end{verbatim}

Generated Answer:
\begin{verbatim}
This function first creates a list of random names, ages....
\end{verbatim}

It is noted that the model writes an explanation rather than continuing the code.

However, the model appears less prone to this error when using randomly initialized embeddings (which succeeded in this test). A potential reason is that the training dataset contains many non-code explanation answers, causing the signal to write natural language to dominate. It is hypothesized that the combination of strong initialization and fine-tuning caused the model to overfit to this specific data distribution.

\subsubsection{DS-1000}

For DS-1000, significantly better functionality was observed in all models when using a heuristic initialization. This is attributed to two primary factors.

First, as a code completion dataset, the provided context prior to generation was significantly longer. This clarified the objective of function completion rather than natural language generation. Furthermore, the inclusion of \texttt{BEGIN SOLUTION} in the code facilitated instruction following. Additionally, the DS-1000 benchmark exhibited the highest density of new tokens, indicating greater similarity to the training data, which likely mitigated overfitting issues.

In this setting, heuristic initialization proved superior for all model versions, as it provides a robust starting point for fine-tuning by appending semantic meaning to embeddings. Regarding the separate experiment of full (sequential) fine-tuning, no significant performance difference was noted, with an equal number of correct answers observed across both benchmarks.

\subsection{LoRA vs Full Finetuning}

When comparing full fine-tuning (where all parameters are unfrozen sequentially to conserve memory), it is observed that LoRA performed better in this setting. This is likely attributable to the model's pre-existing proficiency in Python, where a low-rank approximation suffices for vocabulary expansion. LoRA facilitates fine-tuning while minimizing the catastrophic forgetting often associated with limited datasets and small batch sizes.

In experiments comparing CE and KL training for embeddings, the total number of correct answers was identical across both benchmarks, though performance variance was observed between methods. CE performed better for DS-1000, which aligns more closely with the training data, whereas KL was superior for BigCodeBench, resulting in less model degradation. However, a full comparison with non-LoRA methods would require more comprehensive testing and non-sequential fine-tuning (necessitating higher VRAM). While this small-scale experiment was conducted to detect differences at a granular level, further experimentation is warranted.

Nevertheless, the distinction observed with LoRA was conclusive, suggesting that these results may extend to other scenarios. It is hypothesized that for shorter context code completion, next-token prediction using Cross-Entropy loss suffices. In contrast, for open-ended problems like BigCodeBench, the richer semantic information captured by knowledge distillation renders the proposed method more effective.

\subsection{Challenges}

Training an LLM is inherently complex, and establishing a new pipeline with a novel training objective presented several significant challenges:

\begin{itemize}
\item \textbf{Tokenization:} Extending an existing tokenizer is non-trivial (see Appendix~\ref{sec:voc_met}). Most pre-existing BPE implementations do not support continued training while preserving the exact identity of original tokens. Consequently, the extended vocabulary must be added manually as special tokens, preventing the creation of new merge rules. This lack of merge rules can disrupt standard tokenization, necessitating a meticulous approach to token addition. While this area warrants further exploration, it remained outside the scope of the core research question.

\item \textbf{Preparing dataset:} Precise data preprocessing is critical for instruction training. Token alignment must be exact, as a single indexing error can shift the sequence and disrupt training entirely. The methodology employed is detailed in Appendix~\ref{sec:voc_map}. Since there is no straightforward method for this alignment, empirical verification is required to map special tokens to answers across different tokenizations. Given the complexity, multiple iterations were required at every stage of preprocessing and training to ensure stability.

\item \textbf{Building training pipeline:} unlike standard LLM training where code is often provided by labs (e.g., DeepSeek), implementing a novel training objective requires building infrastructure from scratch. In phase 1, new embeddings were trained separately, which necessitated modifying the model architecture to allow the partitioning of embedding layers—a functionality not typically supported. While the current code serves research purposes, the development of an open-source, HuggingFace-compatible version would greatly enhance accessibility and facilitate future experimentation.

\item \textbf{Setting up a fair experiment:} The experimental design aimed to isolate variables while ensuring all methods were evaluated optimally based on current literature. Identical hyperparameters were maintained across all trials, avoiding method-specific tuning or early stopping. Although this uniform approach may theoretically limit individual peak performance, it ensures that observed differences are attributable solely to the distinct core training objectives rather than hyperparameter variance.

\item \textbf{Hardware and compute:} The computational demands of LLM training were mitigated by the focus on vocabulary expansion with limited data, allowing the model to fit on a single GPU without sharding. However, full fine-tuning necessitated training only half the model at a time due to VRAM constraints. Efficiency was further optimized through a custom implementation where a single model instance was loaded to serve as both student and teacher. Since weights are shared (except for new embeddings), distinct tokenizations and pipeline settings were applied in forward passes before the backward pass. This reduced memory usage significantly, albeit at the cost of increased implementation complexity.
\end{itemize}

\section{Conclusion}
This section summarizes the performance of the proposed method, discusses key takeaways regarding training dynamics, and outlines avenues for future research.

\subsection{Performance}

Results from the benchmarks demonstrate that this novel training method holds significant potential for vocabulary expansion in LLMs. The proposed method outperformed the pre-trained DeepSeek base model, all other vocabulary expansion models, and the fine-tuned base model (without expansion) by a statistically significant factor. This confirms that the mathematical framework not only holds theoretically but is also implementable in deep LLMs.

The core innovation lies in the adaptation of student-teacher training to accommodate differing tokenizations. The primary challenge involves output logit tensors differing in both sequence length and output dimension. The proposed method resolves the sequence length discrepancy by matching identical tokens, while the output dimension mismatch is handled by omitting the extended segment where differences occur. Although the concept (Algorithm \ref{alg:embed-init}) appears straightforward, the engineering required to manipulate tensors is non-trivial. Despite this being a first implementation, the method's superior performance suggests significant potential for further optimization.

\subsection{Takeaways}

It is observed that loss serves as a poor proxy for LLM performance and functionality, particularly during fine-tuning for vocabulary expansion on data similar to pre-training. Since loss is calculated across the entire sequence, the distribution of loss between old and new tokens remains opaque, making the direct correlation between loss and functionality unclear. Future experimentation with weighted loss for different tokens could provide further insight.

Additionally, initialization and training objectives are found to be critical factors; different initializations yielded significantly distinct results rather than negligible variations. It is also noted that the evaluation setup heavily influences outcomes, necessitating extensive experimentation to align the evaluation dataset with the model. The robustness of the proposed method across different settings—unlike other methods which faltered—demonstrates its stability.

Ultimately, the key motivation was the preservation of model knowledge. The method's robustness indicates success in this core objective, effectively avoiding both catastrophic forgetting and overfitting.

\subsection{Future Work}

This novel approach to vocabulary expansion opens several avenues for further experimentation:

\begin{itemize}

\item \textbf{Training embeddings with a mixture of losses:} Currently, input embeddings are trained using KL-divergence, while output embeddings/heads use Cross-Entropy (CE). It is proposed that a mixture of KL and CE for input embeddings could better capture semantic information while aiding next-token prediction. Given the observed sparsity of embeddings, conflict between loss functions is likely minimal. This hybrid approach could enhance robustness in scenarios with a high density of new tokens (where the KL signal is weaker) or divergent sequences. Since the current method relies on surrounding context to infer values for trainable tokens, the signal weakens as the proportion of new tokens increases; a mixed loss could mitigate this dependency.

\item \textbf{Temperature scaling:} In conjunction with cross-entropy, temperature scaling at the softmax layer warrants investigation, as it can profoundly impact student-teacher training \cite{hinton2015distilling, zheng2024knowledge}.

$$\text{softmax}(z_i/T) = \frac{\exp(z_i/T)}{\sum_{j=1}^{K} \exp(z_j/T)}$$

where $T$ is the temperature parameter controlling the softness of the probability distribution.

The current implementation was relatively naive, lacking explicit optimization for student-teacher training. It is hypothesized that increasing the temperature, combined with a cross-entropy loss component, could further improve training. This addition would support next-token prediction (addressing failures seen in randomly initialized models on BigCodeBench) while amplifying signals from other probable tokens learned during pre-training.

\item \textbf{Mechanistic interpretation:} Further experiments could explore mechanistic interpretation using different metrics (e.g., cosine similarity, MSE) or by evaluating distinct activations and hidden states. This thesis focused primarily on evaluating the values of the trained embeddings themselves.

\item \textbf{Different initialization:} While mean initialization was used for input embeddings and the first token composite embedding for the head, alternative linear combinations could be explored to potentially enhance performance.

\item \textbf{Tokenization strategy:} Tokens were manually added to the existing tokenizer for this experiment. Future work could improve extended tokenization by continuing training from the existing tokenizer, resulting in a denser tokenization closer to conventional BPE. This would likely enhance performance, particularly when combined with the proposed fine-tuning method.

\item \textbf{Training on other data types, larger models, and full fine-tuning:} Of particular interest is scaling this method to larger models and comprehensive full fine-tuning on novel data types, rather than data similar to the pre-training set.
\end{itemize}

\newpage

\printbibliography
\newpage

\appendix

\section{Mechanistic Interpretation}
\label{sec:mechanistic}

\subsection{Embeddings Direction in Vector Space}

In the process of training models for the purposes of downstream performance it was discovered that the embeddings direction in vector space was heavily dependent on if one used cross-entropy loss or KL-divergence loss. The motivation to conduct this analysis to begin with was to investigate to what extent heuristic initializations of embeddings are justified. To this end the following analysis was done after Phase 1:

\begin{enumerate}
\item Tokenize all new tokens using original tokenizer, example: new token [eratosthenes] $\rightarrow$ [era, tos, the, nes]
\item Look up embedding for each [$e_{eratosthenes}$] $\rightarrow$ [$e_{era}$, $e_{tos}$, $e_{the}$, $e_{nes}$]
\item Compare cosine similarity for first, average of intermediates (if there are 3 or more subtokens) and last token.
\end{enumerate}

Cosine similarity is defined as:
$$\text{Cosine similarity} = \frac{\mathbf{A} \cdot \mathbf{B}}{||\mathbf{A}|| \cdot ||\mathbf{B}||} = \frac{\sum_{i=1}^{n} A_i B_i}{\sqrt{\sum_{i=1}^{n} A_i^2} \sqrt{\sum_{i=1}^{n} B_i^2}}$$

In simple terms it shows how similar the angle is in vector space, without taking magnitude into consideration. A high cosine similarity of 1 means that they point in the same direction in vector space, 0 are orthogonal, and -1 opposite direction.

\begin{figure}[htbp]
  \centering
  %============================
  % Wrap in a zero‐width box so that negative hspace moves it past textwidth:
  %============================
  \makebox[0pt][l]{%

    \hspace*{-10.2cm}%   % Shift the entire TikZ picture 1cm to the left
    \begin{tikzpicture}
      %--------------------------------------------------
      % First subplot: KL Optimized Embeddings
      %--------------------------------------------------
      \begin{axis}[
          name=plot1,
          xlabel=Epoch,
          ylabel=Similarity of new embedding to its composites,
          title={KL Optimized Embeddings},
          width=0.6\textwidth,
          height=8cm,
          grid=major,
          xmin=0,
          xmax=12,
          cycle list={
            {red,   solid, mark=x, mark options={solid}},
            {blue,  solid, mark=x, mark options={solid}},
            {green, solid, mark=x, mark options={solid}}
          }
        ]
        % First Token (red with × markers)
        \addplot coordinates {
          (0,0.00056535634) (1,0.01670545) (2,0.046268243)
          (3,0.061200596) (4,0.06725986) (5,0.06972933)
          (6,0.07052365) (7,0.07071451) (8,0.07075828)
          (9,0.07137306) (10,0.07191613) (11,0.071994826)
          (12,0.07198474)
        };
        % Intermediate Tokens (blue with × markers)
        \addplot coordinates {
          (0,0.0010031206585498977) (1,0.007886484065280435)
          (2,0.013098017453087559) (3,0.016131679231295555)
          (4,0.018447284189309795) (5,0.019249914715805186)
          (6,0.019305175602511) (7,0.019518090382582325)
          (8,0.019539272084821487) (9,0.020025659858345472)
          (10,0.01997504821823289) (11,0.019996132609305967)
          (12,0.020009008139885705)
        };
        % Last Token (green with × markers)
        \addplot coordinates {
          (0,0.00076339365) (1,0.009277457) (2,0.020059114)
          (3,0.025474334) (4,0.028253485) (5,0.029328804)
          (6,0.029791417) (7,0.030028269) (8,0.030066615)
          (9,0.030582085) (10,0.03108529) (11,0.031227667)
          (12,0.031268325)
        };
      \end{axis}

      %--------------------------------------------------
      % Second subplot: CE Optimized Embeddings
      %--------------------------------------------------
      \begin{axis}[
          name=plot2,
          at={(plot1.outer east)},
          anchor=outer west,
          xlabel=Epoch,
          xshift=-0.2cm,
          title={CE Optimized Embeddings},
          width=0.6\textwidth,
          height=8cm,
          grid=major,
          xmin=0,
          xmax=12,
          cycle list={
            {red,   solid, mark=x, mark options={solid}},
            {blue,  solid, mark=x, mark options={solid}},
            {green, solid, mark=x, mark options={solid}}
          }
        ]
        % First Token (red with × markers)
        \addplot coordinates {
          (0,0.00056535634) (1,0.0027978697) (2,0.009529331)
          (3,0.014691683) (4,0.016751116) (5,0.018077236)
          (6,0.0184654)   (7,0.018612541) (8,0.018627556)
          (9,0.019006161) (10,0.01927732) (11,0.019369798)
          (12,0.01937251)
        };
        % Intermediate Tokens (blue with × markers)
        \addplot coordinates {
          (0,0.0010031206585498977) (1,0.005274525136580148)
          (2,0.005685454156870644) (3,0.0057796715804236546)
          (4,0.005642794416856085) (5,0.006295786827639409)
          (6,0.006509009814264531) (7,0.006609379952832328)
          (8,0.006619187025085691) (9,0.006766016672006755)
          (10,0.006754113682050622) (11,0.006823423310536249)
          (12,0.006826085981179628)
        };
        % Last Token (green with × markers)
        \addplot coordinates {
          (0,0.00076339365) (1,0.008078774) (2,0.01561947)
          (3,0.019558191) (4,0.021674216) (5,0.022784853)
          (6,0.023322973) (7,0.023507671) (8,0.02353283)
          (9,0.02385881)  (10,0.0242174) (11,0.024310885)
          (12,0.02432322)
        };
      \end{axis}

      %--------------------------------------------------
      % Manual legend to the right of plot2
      %--------------------------------------------------
      \node[anchor=north west] at ([xshift=0cm]plot2.outer east) {
        \begin{tabular}{l}
          \textcolor{red}{\rule{0.5cm}{1pt}} First Token \\
          \textcolor{blue}{\rule{0.5cm}{1pt}} Intermediate(s)  \\
          \textcolor{green}{\rule{0.5cm}{1pt}} Last Token \\
        \end{tabular}
      };
    \end{tikzpicture}%
  }
  \caption{Average cosine similarity between new token embeddings and their composite token embeddings for KL optimized embeddings and CE optimized embeddings.}
  \label{fig:embedding_similarity}
\end{figure}

The result here is really interesting, where we see that for randomly initialized embeddings that start off almost orthogonal, the training objective alters what composite embedding they approach the most in similarity.

KL-divergence loss from teacher makes embeddings more aligned with first token embedding, so $e_{eratosthenes}$ would be most similar to $e_{era}$ in this setting. Intuitively, this makes sense since we are optimizing that embedding to produce similar probability distribution as pretrained setting, and in this setting the first token has the most context. For example if we have predicted token 'era' the probability for the token 'tos' is almost a certainty given context.

Similarly, when training for CE loss (next token prediction) we optimize that embedding to predict next token after the word 'eratosthenes'. Since the model as pretrained is trained using next token prediction, it's natural that the optimization would lead to it aligning more with the last embedding corresponding to the last token when eratosthenes is tokenized using old tokenizer.

This result is fascinating and gives a novel interpretation of how LLMs interpret embeddings.

\subsection{Head}

Similarly, we can analyze the rows of head of LLM, otherwise called output embeddings.

\begin{figure}[htbp]
  \centering
  %============================
  % Wrap in a zero‐width box so that negative hspace moves it past textwidth:
  %============================
  \makebox[0pt][l]{%
    \hspace*{-10.3cm}%   % Shift the entire TikZ picture 1cm to the left
    \begin{tikzpicture}
      %--------------------------------------------------
      % First subplot: CE Optimized Head
      %--------------------------------------------------
      \begin{axis}[
          name=plot1,
          xlabel=Epoch,
          ylabel=Similarity of new embedding to its composites,
          title={CE Optimized Head},
          width=0.55\textwidth,
          height=8cm,
          grid=major,
          xmin=0,
          xmax=12,
          cycle list={
            {red,   solid, mark=x, mark options={solid}},
            {blue,  solid, mark=x, mark options={solid}},
            {green, solid, mark=x, mark options={solid}}
          }
        ]
        % First Token (red with × markers)
        \addplot coordinates {
          (0,0.00017873052) (1,0.03775822) (2,0.07361488)
          (3,0.089767955) (4,0.09729063) (5,0.100483075)
          (6,0.101578705) (7,0.1018784) (8,0.101905726)
          (9,0.10446524) (10,0.1060419) (11,0.10643549)
          (12,0.10646778)
        };
        % Intermediate Tokens (blue with × markers)
        \addplot coordinates {
          (0,-0.00020895857195337783) (1,0.008932775616509995)
          (2,0.015117165708985318) (3,0.01740408577854563)
          (4,0.01821888561411386) (5,0.01850627009668548)
          (6,0.018507341482615014) (7,0.018485073459163393)
          (8,0.01847073245420918) (9,0.01888068056539579)
          (10,0.01905157277594727) (11,0.019072461687630556)
          (12,0.01906286753231028)
        };
        % Last Token (green with × markers)
        \addplot coordinates {
          (0,-0.0011045927) (1,0.011135558) (2,0.020283313)
          (3,0.024099207) (4,0.025771985) (5,0.026386071)
          (6,0.026540367) (7,0.026555186) (8,0.026548065)
          (9,0.027239077) (10,0.027633661) (11,0.02769506)
          (12,0.027690062)
        };
      \end{axis}

      %--------------------------------------------------
      % Second subplot: KL+CE Optimized Head
      %--------------------------------------------------
      \begin{axis}[
          name=plot2,
          at={(plot1.outer east)},
          anchor=outer west,
          xshift=-0.2cm,
          xlabel=Epoch,
          title={CE Optimized Head on KL optimized embeddings},
          width=0.55\textwidth,
          height=8cm,
          grid=major,
          xmin=0,
          xmax=12,
          cycle list={
            {red,   solid, mark=x, mark options={solid}},
            {blue,  solid, mark=x, mark options={solid}},
            {green, solid, mark=x, mark options={solid}}
          }
        ]
        % First Token (red with × markers)
        \addplot coordinates {
          (0,0.00017873052) (1,0.039489135) (2,0.08224825)
          (3,0.103158414) (4,0.11302103) (5,0.117454365)
          (6,0.11922648) (7,0.11972436) (8,0.119781286)
          (9,0.1228162) (10,0.12479725) (11,0.125358)
          (12,0.12541807)
        };
        % Intermediate Tokens (blue with × markers)
        \addplot coordinates {
          (0,-0.00020895857195337783) (1,0.010807843049071546)
          (2,0.022526730789974247) (3,0.02862519458211937)
          (4,0.03146885121415166) (5,0.03277591149518052)
          (6,0.0332995510537006) (7,0.033438342628765036)
          (8,0.033455082812164534) (9,0.03428741389264663)
          (10,0.034853121974965796) (11,0.03502108164174461)
          (12,0.03503836907483852)
        };
        % Last Token (green with × markers)
        \addplot coordinates {
          (0,-0.0011045927) (1,0.012915728) (2,0.026944151)
          (3,0.033410598) (4,0.036480352) (5,0.037750684)
          (6,0.038232833) (7,0.038355503) (8,0.038369346)
          (9,0.039234683) (10,0.039787646) (11,0.039928555)
          (12,0.03994147)
        };
      \end{axis}

      %--------------------------------------------------
      % Manual legend to the right of plot2
      %--------------------------------------------------
      \node[anchor=north west] at ([xshift=-0.5cm]plot2.outer east) {
        \begin{tabular}{l}
          \textcolor{red}{\rule{0.5cm}{1pt}} First Token \\
          \textcolor{blue}{\rule{0.5cm}{1pt}} Intermediate(s)  \\
          \textcolor{green}{\rule{0.5cm}{1pt}} Last Token \\
        \end{tabular}
      };
    \end{tikzpicture}%
  }
  \caption{Average cosine similarity between new head embeddings and their composite token embeddings for CE optimized head and KL+CE optimized head.}
  \label{fig:head_similarity}
\end{figure}

Here, the output embeddings come to most align with that of first composite token-embedding. This is completely expected behavior considering the pretrained setting. If we are in old pretrained setting predicting tokens [num, py], and new token is [numpy], the output embedding for [numpy] will approach that of [num]. The token [py] is never being used with our new tokenization, and subsequently, the output embedding does not align itself with it as much despite it being a composite embedding of numpy. 

\subsection{Conclusion}

From this analysis there are several key takeaways.

Firstly, we note that when we optimize both input and output embeddings as seen in Figure \ref{fig:embedding_similarity} and \ref{fig:head_similarity}, we observe that the cosine similarity on average is lower with intermediate embeddings which indicates that the first and last tokens for whole words encode more relevant information than intermediate ones. 

Secondly, when we optimize an embedding for next token prediction, it favors the ``direction'' of the last composite embedding. So in the case $[e_{\text{numpy}}] \rightarrow [e_{\text{num}}, e_{\text{py}}]$, we get that $e_{\text{numpy}}$ is more similar to $e_{\text{py}}$. A hypothesis for why this is the case is that individual indices on the embedding vector of $e_{\text{py}}$ are key to predict the next correct token in sequence. On the other hand, when we optimize for KL-divergence loss we get similar cosine similarity between optimized token and the last token (around 0.03), in addition to a high average to the token-embedding ($e_{\text{num}}$ in this case). This would suggest that for a word, the first token of that word on average affects the whole probability distribution for the next word compared to the last token that more so affects the value of the logits for a very small subset of the probability space.

The motivation for this is two-fold:
\begin{itemize}
    \item Looking at Figure \ref{fig:phase1_kl_loss}, we can see that training with the cross entropy objective for embeddings by and large is disconnected from KL-divergence loss, and in the case of mean initialization even increases KL-divergence loss. This in essence loses nuance of what could be the next probable token.
    \item Figure \ref{fig:embedding_similarity} shows that when we train for KL-divergence loss, the optimized embedding comes to more align with that of the first composite token-embedding. This would indicate that the first token in a word encodes more information about what is the context by and large and what could be a probable subsequent token.
\end{itemize}

Finally, for both input and output embedding, the cosine similarity is quite low after training when starting from a random initialization ($<0.1$). However, the losses as seen in Figure \ref{fig:phase1_ce_loss} and \ref{fig:phase1_kl_loss} decrease substantially. This points to embeddings not encoding information by being in a specific direction in vector space but rather because a few individual indices or combination of indices encodes most if not all information about individual tokens. In other words, if we were to view the model dimension as a vector space, it's only a small subspace that is relevant and that differs for each token. The graphs here only show cosine similairy from when we used random initialization, which lead them to start of almost orthogonal to our pretrained embeddings, but it was also found that when when we started with an average initialization and got a similarity of around 0.6 and then trained, the cosine similarity was almost unchanged despite loss decreasing as seen in Figure \ref{fig:phase1_kl_loss}. 

\newpage

\section{Vocabulary Expansion Method}\label{sec:voc_met}

To expand the vocabulary of a pre-trained tokenizer, we developed a systematic approach that identifies and incorporates the most relevant tokens from our training data. This method ensures that we only add tokens that would genuinely improve tokenization efficiency without disrupting the existing token structure.

\begin{algorithm}
  \caption{Vocabulary Expansion for Tokenizer}
  \label{alg:vocab-expansion}
  \begin{algorithmic}[1]
    \Require Original tokenizer $T$, training data $D$, number of tokens to add $N$
    \Ensure Extended tokenizer $T'$ with $N$ additional tokens
    \State Collect training data $D$ relevant to the target domain
    \State Train a new BPE tokenizer $T_{\mathrm{new}}$ from scratch on $D$
    \State Tokenize $D$ with $T_{\mathrm{new}}$ to identify token occurrences
    \State Count occurrences of each token in $D$
    \State Remove tokens that are just numbers (e.g., “12”, “31”)
    \State Remove tokens that are substrings of tokens in the original tokenizer $T$
    \State Remove duplicate tokens already present in $T$
    \State Sort remaining tokens by occurrence count in descending order
    \State Select the top $N$ most frequently occurring tokens $\{t_1,\dots,t_N\}$
    \State Add selected tokens to the original tokenizer: $T' \leftarrow T \cup \{t_1,\dots,t_N\}$
    \Return Extended tokenizer $T'$
  \end{algorithmic}
\end{algorithm}

This algorithm allows us to systematically identify and incorporate the most valuable tokens from our domain-specific training data, improving the tokenizer's efficiency for our specific use case without disrupting its existing structure.

\newpage

\section{Token Sequence Mapping}\label{sec:voc_map}

When comparing how original and extended tokenizers process the same text, we need to identify where they match and where they diverge. The created mappings are used to pairwise compare logits that equate to same place in text.

\begin{algorithm}[H]
\caption{Token Sequence Mapping Algorithm}
\label{alg:token-mapping}
\begin{algorithmic}[1]
  \Require Text $T$, original tokenizer $T_{\mathrm{orig}}$, extended tokenizer $T_{\mathrm{ext}}$
  \Ensure Mappings between original and extended tokenizations
  \State Tokenize $T$ with $T_{\mathrm{orig}}$ into sequence $O$ and with $T_{\mathrm{ext}}$ into sequence $E$
  \State Initialize indices $i \gets 1$, $j \gets 1$
  \While{$i \le |O|$ \textbf{and} $j \le |E|$}
    \If{$O_i = E_j$}
      \State Record matching pair $(O_i, E_j)$
      \State $i \gets i + 1$, \quad $j \gets j + 1$
    \Else
      \State $i_{\mathrm{start}} \gets i$, \quad $j_{\mathrm{start}} \gets j$
      \State Look ahead up to 20 tokens to find the smallest offsets $d,d'$ such that $O_{i+d} = E_{j+d'}$
      \If{resynchronization point $(i+d,j+d')$ found}
        \State Record divergent groups $O_{i_{\mathrm{start}}:i+d-1}$ and $E_{j_{\mathrm{start}}:j+d'-1}$
        \State $i \gets i + d$, \quad $j \gets j + d'$
      \Else
        \State Record divergent pair $(O_i, E_j)$
        \State $i \gets i + 1$, \quad $j \gets j + 1$
      \EndIf
    \EndIf
  \EndWhile
  \If{$i \le |O|$}
    \State Record remaining $O_{i:|O|}$ as divergent
  \EndIf
  \If{$j \le |E|$}
    \State Record remaining $E_{j:|E|}$ as divergent
  \EndIf
  \Return Complete mapping of matching and divergent sequences
\end{algorithmic}
\end{algorithm}

Consider an example of the code string when having different tokenization: \newline
\texttt{int CellConnectDLS(void) \{ return 0; \}}

Original tokenization: \newline
[\texttt{"int"}, \texttt{" "}, \texttt{"Cell"}, \texttt{"Connect"}, \texttt{"D"}, \texttt{"L"}, \texttt{"S"}, \texttt{"("}, \texttt{"void"}, \texttt{")"}, \texttt{" \{"}, \texttt{" "}, \texttt{"return"}, \texttt{" "}, \texttt{"0"}, \texttt{";"}, \texttt{" \}"} ]

Extended tokenization: \newline
[\texttt{"int"}, \texttt{" "}, \texttt{"CellConnect"}, \texttt{"DLS"}, \texttt{"("}, \texttt{"void"}, \texttt{")"}, \texttt{" \{"}, \texttt{" "}, \texttt{"return"}, \texttt{" "}, \texttt{"0"}, \texttt{";"}, \texttt{" \}"} ]

The algorithm would output:
\begin{verbatim}
Similar mappings: 
[(0, 0), (1, 1), (7, 4), (8, 5), (9, 6), (10, 7), (11, 8), (12, 9), 
(13, 10), (14, 11), (15, 12), (16, 13)]

Divergent mappings: 
[([2, 3, 4, 5, 6], [2, 3])]  # "CellConnectDLS" tokenized differently
\end{verbatim}

This example shows how the extended tokenizer processes the function using 14 tokens instead of 17, with the specialized domain tokens \texttt{"CellConnect"} and \texttt{"DLS"} replacing five tokens in the original tokenization. These mappings are used in training to compare logits that equate to same plces in text.

\section{Hardware \& Cost}
\label{sec:hardware}

As mentioned in Discussion, much compute was spent on experimenting and therefore did not directly contribute to the end result. However, if compute were to be estimated, it was approximately one GPU-month on an L40S, which would amount to approximately 40,000 SEK on a single GPU AWS EC2 instance.

A more efficient way of conducting experiments if done again and/or at larger scale would be to have a multi-GPU setup (ideally with more VRAM than 46GB per GPU) and run experiments and ablations in parallel.

% ---------- REFERENCES ----------

\end{document}